%% file: bare_jrnl.tex
\documentclass[journal]{IEEEtran}

\usepackage{amsmath,amssymb,amsfonts}
\usepackage{amsmath,array, amscd}
\usepackage{bbm}
\usepackage{algorithmic}
\usepackage{graphicx}
\usepackage{textcomp}
\usepackage{algorithm}
\usepackage{multirow}
\usepackage{todonotes}
\usepackage{booktabs}
\usepackage{verbatim}
\usepackage[symbol]{footmisc}
\usepackage{titlesec}

\usepackage[noadjust]{cite}

\renewcommand{\thefootnote}{\fnsymbol{footnote}}

\newcommand\blfootnote[1]{%
  \begingroup
  \renewcommand\thefootnote{}\footnote{#1}%
  \addtocounter{footnote}{-1}%
  \endgroup
}

\usepackage[colorlinks=true, citecolor=black,linkcolor=black,urlcolor=blue]{hyperref}

\ifCLASSINFOpdf
\else
\fi
\begin{document}
\title{Robust Prototypical Few-Shot Organ Segmentation with Regularized Neural-ODEs}

\author{
Prashant Pandey, Mustafa Chasmai, Tanuj Sur, Brejesh Lall
}


\markboth{Journal of \LaTeX\ Class Files,~Vol.~14, No.~8, August~2015}%
{Shell \MakeLowercase{\textit{et al.}}: Robust Prototypical Few-Shot Organ Segmentation with Regularized Neural-ODEs}



\maketitle

\begin{abstract}
Despite the tremendous progress made by deep learning models in image semantic segmentation, they typically require large annotated examples, and increasing attention is being diverted to problem settings like Few-Shot Learning (FSL) where only a small amount of annotation is needed for generalisation to novel classes. This is especially seen in medical domains where dense pixel-level annotations are expensive to obtain. In this paper, we propose Regularized Prototypical Neural Ordinary Differential Equation (R-PNODE), a method that leverages intrinsic properties of Neural-ODEs, assisted and enhanced by additional \textit{cluster} and \textit{consistency} losses to perform Few-Shot Segmentation (FSS) of organs. R-PNODE constrains support and query features from the same classes to lie closer in the representation space thereby improving the performance over the existing Convolutional Neural Network (CNN) based FSS methods. We further demonstrate that while many existing Deep CNN-based methods tend to be extremely vulnerable to adversarial attacks, R-PNODE exhibits increased adversarial robustness for a wide array of these attacks. We experiment with three publicly available multi-organ segmentation datasets in both in-domain and cross-domain FSS settings to demonstrate the efficacy of our method. In addition, we perform experiments with seven commonly used adversarial attacks in various settings to demonstrate R-PNODE's robustness. R-PNODE outperforms the baselines for FSS by significant margins and also shows superior performance for a wide array of attacks varying in intensity and design.
\blfootnote
{Copyright (c) 2019 IEEE. Personal use of this material is permitted. However, permission to use this material for any other purposes must be obtained from the IEEE by sending a request to pubs-permissions@ieee.org. 

Prashant Pandey, Mustafa Chasmai and Brejesh Lall are with the Department of Electrical Engineering and Department of Computer Science, Indian Institute of Technology Delhi, New Delhi 110016, India. Tanuj Sur is with Chennai Mathematical Institute, India. The first two authors contributed equally to this work. Email: getprashant57@gmail.com, mustchasmai@gmail.com, surtantheta@gmail.com,
brejesh@ee.iitd.ac.in.
}
% \blfootnote{*Equal contribution}
\end{abstract}
\begin{IEEEkeywords}
Few-shot Segmentation, Neural-ODEs, Adversarial Robustness, Medical Image Segmentation.
\end{IEEEkeywords}
\section{Introduction}
\label{sec:introduction}
Deep learning-based methods are widely successful~\cite{guo2018review, wang2018understanding} in image segmentation having dense pixel-level annotations. The fully-supervised methods perform extremely well with a large number of annotated examples. It may be quite straightforward for existing approaches to employ techniques like transfer learning and fine-tuning when well-organized large-scale datasets with natural images are easily available. This is not the case for medical image segmentation. In medical domains~\cite{gut2022benchmarking}, it is not always feasible to collect dense pixel-level annotations, specifically when the medical images vary in image features or characteristics depending upon the location from where the images are collected, camera characteristics, modality (CT, MRI, X-Ray etc.) and the task (type of organ to segment) at hand. 

While for some organs, the 3D scans may be available in abundance for analysis and learning, a few body organs need sophisticated costly medical devices to capture them thereby rendering the fully-supervised approaches mostly ineffective due to over-fitting with abundantly labelled examples and poor generalization on scarce classes/organs. \textcolor{black}{Also, fully supervised methods would need to directly rely on a significant amount of annotations of these scarce or novel target organs to be able to perform well.} Even transfer learning or fine-tuning methods won't be effective due to the unavailability of \textit{generic} large-scale medical datasets and the presence of significant domain shifts between base/seen classes and novel/unseen classes.
\begin{figure}[!h]
\centering
\includegraphics[width=0.8\linewidth]{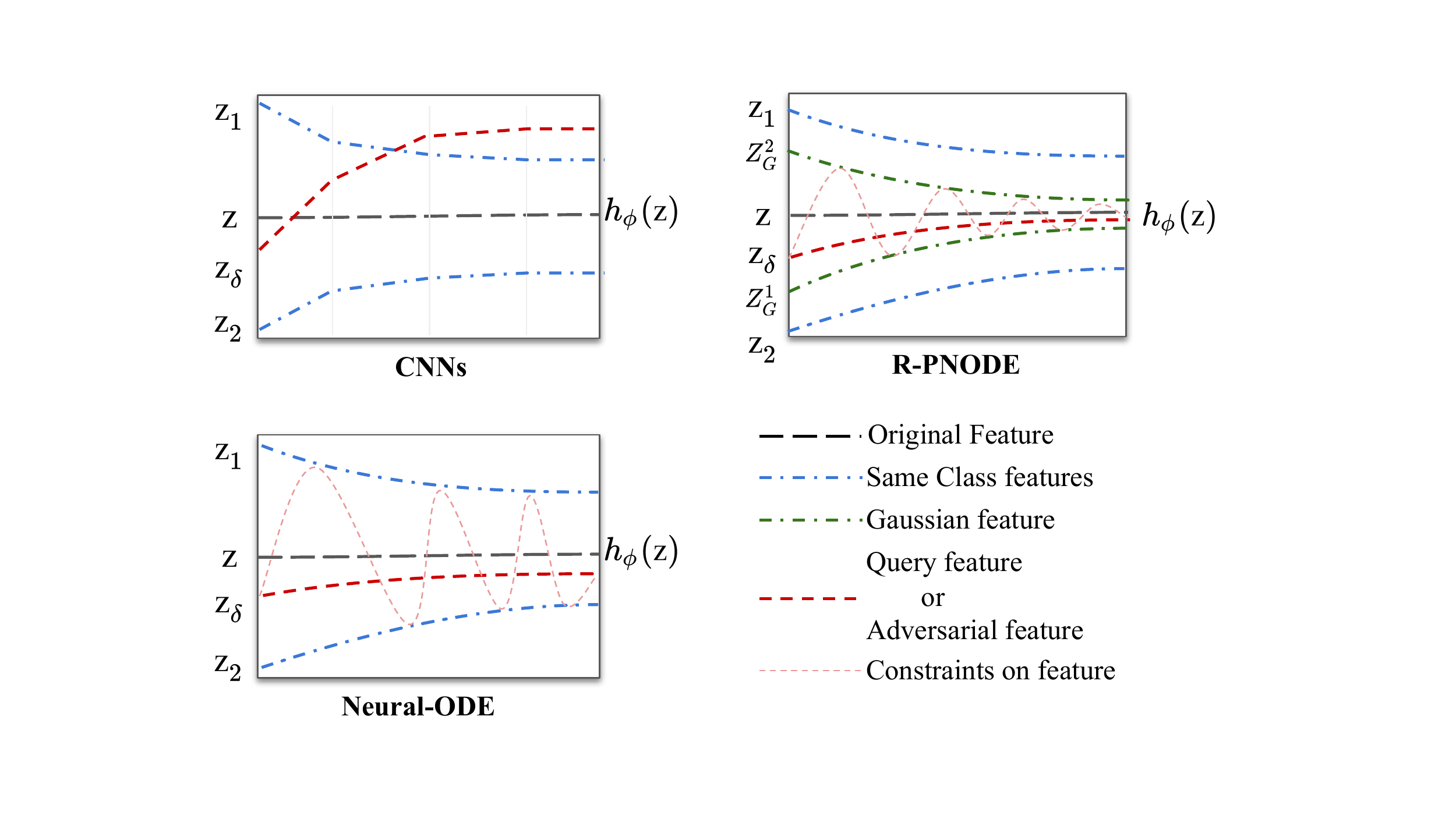}
% \caption{Given an input image and its corresponding Gaussian perturbed examples, the feature representations from the feature extractor $f_{\theta}$ are $Z$, $Z_{1}$, and $Z_{2}$, respectively. Now, during testing, an adversarially perturbed sample with features $Z_{\delta}$ is introduced. The non-intersecting property of integral curves in Neural ODEs imposes intrinsic constraints on the feature representations, so that $Z_{\delta}$ in subsequent layers is sandwiched between $Z_{1}$ and $Z_{2}$. The proposed R-PNODE, further improves this robustness by employing additional cluster loss $\mathcal{L}_{CL}$ and consistency loss $\mathcal{L}_{CON}$ to bring $Z_{1}$ and $Z_{2}$ closer to $Z$. }
\caption{\textcolor{black}{Visualisation of the evolution of the features learnt (y-axis) across the layers in a CNN and across time (x-axis) for a Neural-ODE. Features $\mathrm{Z}_1$ and $\mathrm{Z}_2$ belong to the same class as $\mathrm{Z}$. CNNs do not have explicit constraints on the nature of these features, and the output features of a slightly different/perturbed input or adversarial feature ($\mathrm{Z}_{\delta}$) may be very different from the expected output ($h_{\phi}(\mathrm{Z})$) of original feature $\mathrm{Z}$. On the other hand, Neural-ODEs (bottom left) have the constraint that any two of these curves (feature transformations across the layers) may not intersect. Because features of the same class are being brought closer during training, the features of other samples (like $\mathrm{Z}_{\delta}$) are constrained to belong to the same class, resulting in robustness. In R-PNODE (top right), these constraints are further tightened by populating the sample space with additional Gaussian samples $Z^1_{G}$ and $Z^2_{G}$.}}
\label{fig:rpnode_comp}
\end{figure}

% Further, earlier works like [2] and [3] also mention that acquiring datasets with abundant manual labels is often very expensive and time-consuming as it requires experts with many years of clinical experience. As discussed in [4], annotations for organs like adrenal glands and duodenum are scarce and hard to collect. This is also validated in the paper [5] where all manual annotations were made or overseen by a graduate student with several years of experience annotating CT images. So for scarce organs, FSS can be an optimal framework to learn effectively.

% Also, fully supervised methods would need to rely on a significant amount of annotations of these scarce or novel target organs to be able to perform well. On the other hand,  

Few-Shot Segmentation (FSS) comes to the rescue by leveraging approaches like metric learning~\cite{wang2019panet, snell2017prototypical, li2022few, dong2018few} (e.g. Prototypical Learning) to segment novel or rare classes/organs with examples varying from one to five per class. \textcolor{black}{One of the major challenges arising from the implementation of screening and diagnosis programs is the enormous amount of CT images that must be analysed by medical practitioners. Automated systems are intended to make the interpretation of CT images faster and more accurate, thereby improving the cost-effectiveness of the screening program. Advanced computer-aided diagnosis systems (implemented with ML/Deep learning) have the potential to expedite this process and motivate the research community to effectively design such cost-effective automated systems like FSS that have the ability to learn well even with few annotated examples.} An FSS model has the ability to segment novel classes by learning support and query image relationships from a completely disjoint set of base classes with an ample number of \textcolor{black}{annotated} examples \textcolor{black}{that are freely available}. During testing on novel classes, the model uses the learnt metric to segment unseen regions present in the query images with the help of a few labelled support images.

Past few years, many FSS methods on both natural~\cite{wang2019panet, snell2017prototypical} and medical~\cite{li2022few, ouyang2022self, al2021ifss, wang2021few, cui2020unified, feng2021interactive} images are proposed that employ CNN based feature extractors. These methods are devoid of the capability to \textit{explicitly} force prototypes of support images to lie closer to query image features that may lead to poor performance. This is further aggravated in medical domains where the novel/test query images can be slightly different from the images in the base classes due to variations in data modality, texture, tissue appearance and orientation, camera characteristics, colour intensity and size and shape of the target organs. These additional test query image characteristics can be regarded as perturbations of the query features and the learnt metric in prototypical FSS fails to capture these subtle variations.
% Further, the modern-day safety-critical medical systems are vulnerable to various types of threats and attacks that can cause danger to human life. With the penetration of AI, Machine Learning and Deep Learning models in healthcare and medical domains, it is imperative to make such models robust against different kinds of attacks. By design, these models are data-hungry and need a large amount of labeled training examples to improve their performance and generalizability.
% Past studies have shown that it is not always feasible to annotate medical data, especially for segmentation problems due to the huge time and specific skills it needs to do so. 
Further, due to a lack of well-annotated data, these models are vulnerable to several kinds of white and black-box adversarial attacks~\cite{goodfellow2014explaining,kurakin2018adversarial,madry2017towards}. 

\textcolor{black}{
Medical diagnosis is often a safety-critical task, where a small mistake can cost a human life. In \cite{li2020robust}, the authors discuss how the activities of certain startups indicate that deep learning-based medical imaging systems are potentially applicable for clinical diagnosis in the near future. If indeed complete automation is to be achieved here, the entire pipeline should be able to either handle or at least detect any possible malicious attacks. Although still relatively under-explored, there are quite a few studies working on this problem. Some studies work on improving adversarial robustness \cite{liu2020no, xu2022medrdf}, some propose new benchmarking or testing strategies catered for the medical domain \cite{daza2021towards} and some work on detecting these adversarial attacks \cite{li2020robust}. A survey done recently \cite{kaviani2022adversarial} discussed how a lack of a sufficient amount of high-quality images and labelled data in the medical domain is one of the main reasons for the weaknesses of adversarial defence mechanisms. In a few-shot setting, where the amount of labelled data is extremely low, these weaknesses are bound to be prominent.}

ML practitioners employ FSS
%FSL~\cite{fei2006one} 
to learn patterns using well-annotated base classes, and finally to transfer the knowledge to scarcely annotated novel classes. This knowledge transfer is severely impacted in the presence of adversarial attacks when the support and query samples from novel classes are injected with adversarial perturbations and models fail to recognize organs, important clinical landmarks etc., present in the image thereby questioning the credibility of these FSS methods for medical image segmentation.  
%Existing works propose defence mechanisms~\cite{Kolter,Wong} that cater to specific attacks and their generalizability on different kinds of adversarial attacks in the medical segmentation domain is not fully ascertained. 

Common Adversarial Training mechanisms \cite{goodfellow2014explaining, madry2017towards, zhang2019theoretically} require adversarially perturbed examples shown to the model during training.~\cite{xu2021dynamic} introduced standard adversarial training (SAT) procedure for semantic segmentation. These methods do not guarantee defence when the type of attack differs from the adversarially perturbed examples~\cite{zhang2019limitations,park2020effectiveness}, and it is impractical to expose the model to different kinds of adversarial examples during training.
% These methods do not guarantee defence when the type of attack is different from the adversarially perturbed examples~\cite{zhang2019limitations,park2020effectiveness} and it is impractical to expose the model with different kind of adversarial examples during training itself.
Also, \textit{a common method that handles attacks both on support and query examples of novel classes, is non-existent}. To the best of our knowledge, the adversarial attacks on few-shot segmentation (FSS) with Deep Neural models and their defence mechanisms have not yet been explored and the need for such robust models is inevitable. 

To this end, we propose \textbf{R}egularized \textbf{P}rototypical \textbf{N}eural \textbf{O}rdinary \textbf{D}ifferential \textbf{E}quation (R-PNODE), a novel prototypical few-shot segmentation method based on Neural-ODEs \cite{chen2018neuralode} that provides robustness against slight variations in query image features and defence against different kinds of adversarial attacks in different settings. Owing to the fact that the integral curves of Neural-ODEs are non-intersecting, perturbations in the input lead to small changes in the output as opposed to existing FSS models with CNN-based feature extractors where the output is unpredictable. 
% \textit{This enables R-PNODE to force query features to be constrained by the integral curves of support features belonging to the same class as that of the query and hence they lie closer in the representation space}. 
For a \textit{stable} dynamical system, if an example $\textbf{x}$ is slightly perturbed with $\epsilon$ such that the initial state of the example is $\textbf{x}_0= \textbf{x} + \epsilon$, then at time $t \rightarrow \infty,~\textbf{x}_t \rightarrow \textbf{x}$. This property is desirable for robustness against input perturbations, and to help promote stability in R-PNODE, we employ additive and multiplicative perturbations with Gaussian noise \cite{li2019certified,ford2019adversarial, he2019parametric,byun2022effectiveness,qin2021random} that 
% \textbf{a)} result in an \textit{almost sure exponential stability} \cite{mao2007stochastic} of S-PNODE framework \textbf{b)} 
help to learn robust class-wise prototypes and query features. Additionally, we apply \textit{cluster loss} that enforces prototypes perturbed with Gaussian noise and their non-perturbed counterparts to lie closer in the representation space, enabling R-PNODE to be robust against adversarially attacked support examples in novel classes. Similarly, we apply \textit{consistency loss} between predictions of perturbed query features and the ground-truth labels of their corresponding non-perturbed (or clean) query inputs that render R-PNODE robust against various perturbations (adversarial/non-adversarial) of query sets. 

\textit{Thus, R-PNODE forces query features to be constrained by the integral curves of support features belonging to the same class as that of the query and hence they lie closer in the representation space} as shown in Fig. \ref{fig:rpnode_comp}. \textcolor{black}{This explicit constraining of features along with the proposed losses is absent in Neural-ODEs.}
%The integral curves of Neural-ODEs \cite{chen2018neuralode} have been shown to be non-intersecting, which ensures that adversarial perturbations in the input lead to small changes in the output as opposed to existing FSS models where the output is unpredictable.
In this paper, we make the following contributions:
%- We propose a novel adversarial training mechanism for few-shot learning that can work in tandem with PNODE to improve performance and generalizability against adversarial attacks.
% - We experiment by attacking both support and query, and extend the standard adversarial training procedure to handle both attacks simultaneously.
\begin{enumerate}
\item We propose a novel prototypical FSS method, R-PNODE, that leverages Neural-ODEs regularized with cluster and consistency losses to outperform existing few-shot organ segmentation methods by large margins.
\item We extend Standard Adversarial Training
to the FSS domain and handle attacks on both support and query.
\color{black}
\item We demonstrate how our method is adversarially robust, with  the ability to handle adversarial attacks like FGSM~\cite{goodfellow2014explaining}, PGD~\cite{madry2017towards}, BIM\cite{kurakin2018adversarial}, CW \cite{carlini2017towards}, Auto-Attack\cite{croce2020reliable}, DAG~\cite{xie2017adversarial} and SMIA~\cite{qi2021stabilized} differing in intensity and design when applied to support or query image. We show R-PNODE's efficacy by repeating these experiments on three publicly available multi-organ segmentation datasets BCV~\cite{landman2015miccai}, CT-ORG~\cite{rister2020ct} and DECATHLON~\cite{simpson2019large} for both in-domain and cross-domain settings on novel classes.
\end{enumerate}
\color{black}
\section{Related Works}
\subsection{Neural-ODEs:} 
Deep learning models such as ResNets~\cite{he2016deep} learn a sequence
of transformations by mapping input \textbf{x} to output \textbf{y} by composing a sequence of
transformations to a hidden state. In a
ResNet block, computation of a hidden layer representation
can be expressed using the following transformation: %$\mathrm{\textbf{h}(t+1)} = \mathrm{\textbf{h}(t)} + f_{\theta}\mathrm{(\textbf{h}(t), t)}$
\begin{equation}
    \mathrm{\textbf{h}(t+1)} = \mathrm{\textbf{h}(t)} + f_{\theta}\mathrm{(\textbf{h}(t), t)}
    \label{eq:resnet}
\end{equation}

where $\mathrm{t \in \{0, \hdots, T\}}$ and $\mathrm{\textbf{h}}:[0,\infty] \rightarrow \mathbb{R}^n$. 
As the number of layers is increased and smaller
steps are taken, in the limit, the continuous
dynamics of the hidden layers are parameterized using an ordinary differential
equation (ODE)~\cite{chen2018neuralode} specified by a neural network, %$\frac{d\textbf{h}\mathrm{(t)}}{d\mathrm{t}} = f_{\theta}\mathrm{(\textbf{h}(t), t)}$
\begin{equation} \label{eq:2}
    \frac{d\textbf{h}\mathrm{(t)}}{d\mathrm{t}} = f_{\theta}\mathrm{(\textbf{h}(t), t)}
\end{equation}

where $f:\mathbb{R}^n \times [0,\infty] \rightarrow \mathbb{R}^n$
denotes the non-linear trainable layers parameterized by
weights $\mathbf{\theta}$.
These layers define the relation between the input $\mathrm{h(0)}$ and output $\mathrm{h(T)}$, at time $\mathrm{T}>0$,  by providing a solution to the ODE initial value problem at terminal time T. Neural-ODEs
are the continuous equivalent of ResNets where the hidden layers can
be regarded as discrete-time difference equations.
\color{black} The time here is a dummy variable loosely corresponding to the equivalent of layer number in ResNets, though it may not have a physical interpretation. The curve \textbf{h}(t), obtained by solving Eq.~\ref{eq:2}, is termed as an integral curve and denotes the $n$-dimensional features at any intermediate time t. Different initial values \textbf{h}(0) lead to different particular solutions which can be thought of as different integral curves obtained as different inputs are passed through the Neural-ODE. \color{black}
%as shown in equation~\ref{eq:resnet}.\\

\subsection{Intrinsic Robustness of Neural-ODEs: }

\begin{figure}[!h]
\centering
\includegraphics[width=0.41\linewidth]{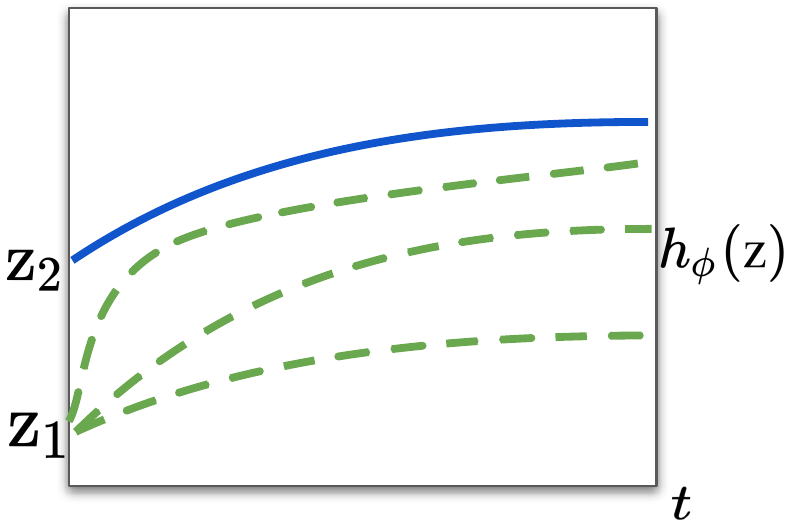}
\includegraphics[width=0.41\linewidth]{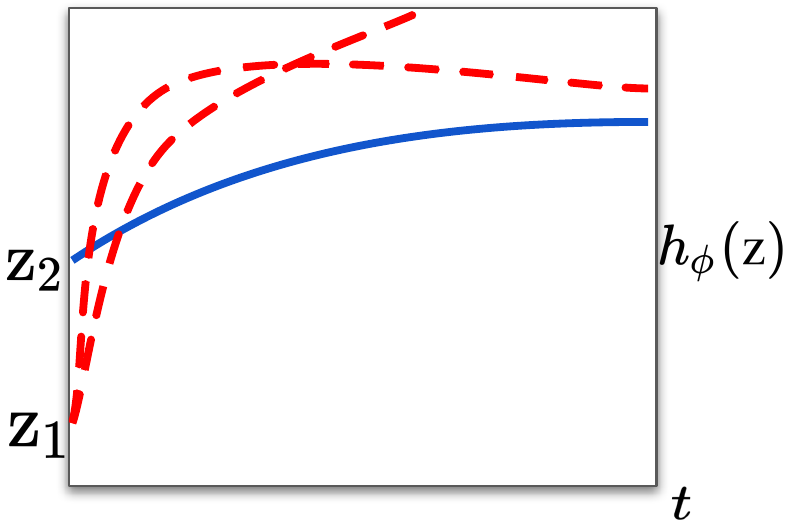}
\caption{
\textcolor{black}{An illustration to explain the non-intersecting property in Neural-ODEs. $\mathrm{z_1}$ and $\mathrm{z_2}$ are two hypothetical data points passed through the Neural-ODE. $h_{\phi}(\mathrm{z})$ (y-axis) is the expected output of any data point $\mathrm{z}$ after time $t$ (x-axis). If the integral curve of $\mathrm{z_2}$ is known to be in the form of the blue curve in the figure, the possible integral curves of $\mathrm{z_1}$ become restricted due to the non-intersecting property. We represent the integral curves that are possible by green dashed lines (left figure) and the ones that are \textit{not} allowed by the non-intersecting property, with red dashed curves (right figure).}}
\label{fig:non_intersect}
\end{figure}

\textcolor{black}{
Multiple works have tried to explain the intrinsic robustness of Neural-ODEs. \cite{chen2018neuralode} argues that the  non-intersecting property of the Neural-ODE's integral curves allows the model to  be intrinsically robust. Because the integral curves are governed by differential equations of the form shown in Eq.~\ref{eq:2}, it follows that if two integral curves intersect, their slopes at the point of intersection must be identical. Following the curves back to the y-axis (Fig.~\ref{fig:non_intersect}), it can be proved that any two intersecting integral curves have to be identical, or equivalently, distinct integral curves cannot intersect. As can be seen in Fig.~\ref{fig:non_intersect}, this leads to restrictions on the possible integral curves and this consequently improves robustness as illustrated in Fig.~\ref{fig:rpnode_comp}. We elaborate on this property, its proof and its consequences on robustness with a supplementary video presentation.
}

\textcolor{black}{
As an alternate theory to account for this intrinsic robustness, \cite{huang2021adversarial} shows that adaptive stepsize ODE solvers commonly used in Neural-ODEs tend to have a gradient-masking effect, and this allows the model to be robust to gradient-based attacks. Multiple studies~\cite{chen2018neuralode,liu2020does,kang2021stable} have applied Neural-ODEs to defend against adversarial attacks.~\cite{yan2019robustness} proposes time-invariant steady Neural-ODE that is more stable than conventional convolutional neural networks (CNNs) in the classification setting.~\cite{kang2021stable} design a Neural-ODE such that the equilibrium points of the ODE solution have Lyapunov-stability, thereby suppressing input perturbations. \cite{liu2020does} explores the equivalents of different regularising noise injection techniques like dropout and gaussian smoothing in Neural-ODEs, and argues how these techniques lead to better, more stable equilibrium points. \cite{jimenez2022lyanet} describe a framework that guarantees exponential convergence and improved adversarial robustness by using a novel Lyapunov loss, which they approximately optimise using a  practical Monte Carlo approach.}

\color{black}

\subsection{Few-shot Segmentation (FSS):} 
% FSL methods seek good generalization and learn transferable knowledge across different tasks with limited data~\cite{fei2006one,zhao2019data,ouyang2019data}. 
%, typically containing
%just a few training samples of the target classes
Few-shot segmentation (FSS)~\cite{roy2020squeeze,wang2019panet,li2020fss,feng2021interactive,ouyang2022self,wang2021few,cui2020unified} aims to perform pixel-level classification for novel classes in a query image when trained on only a few labelled support images. % ~\cite{shaban2017one} proposed a novel technique for one-shot segmentation.
A commonly adopted approach for FSS is based on prototypical networks~\cite{snell2017prototypical,dong2018few,wang2019panet,tang2021recurrent} that employ prototypes to represent typical information for foreground objects present in the support images. 
The prototype is subsequently compared with the query images pixel-wise via cosine similarity, leading to the segmentation mask. 
%Deep learning has become an integral part of the medical imaging community
%for a varied range of tasks like classification, segmentation, detection, etc. Especially due to lack of labelled data, application of few-shot learning has gained a lot of momentum in this domain.
%In few-shot medical image segmentation, most works generate new training data to increase the size of the training set given only a few labels~\cite{zhao2019data,ouyang2019data}. 
%However, whenever a new class needs to be segmented, we need to retrain the model which makes these models difficult to generalize to unseen classes.
In addition to the prototype-based setting, \cite{roy2020squeeze} incorporates `squeeze \& excite’ blocks that avoid the need for pre-trained models for medical image segmentation. \cite{li2020fss} uses a relation network and introduced the FSS-1000 dataset that is significantly smaller as compared to contemporary large-scale datasets for FSS. The data-scarce medical domain has a unique requirement for specialized FSS algorithms \cite{sun2022few,tang2021recurrent}. One common problem setting in this  domain is that  of  organ segmentation, with openly  available organ segmentation datasets \cite{landman2015miccai,rister2020ct,simpson2019large} and specialized methods \cite{kim2021bidirectional,li2022few} addressing their  unique  challenges. \cite{kim2021bidirectional} used a bidirectional gated recurrent unit (GRU) to capture relationships of features across slices. 
\cite{li2022few} uses an interclass segmentation network for organ segmentation on a multi-institution dataset from prostate cancer patients.  Many other methods \cite{ouyang2020self,zhang2021few,mondal2018few} addressing different unique challenges can be found in literature, but we focus more on the commonly used prototypical networks. MetaNODE \cite{zhang2022metanode} was one of the first works to use Neural-ODEs in the Few-shot domain.  
Our work is quite different to MetaNODE as the latter addresses prototype bias reduction. Also, the core technique varies in these two cases. MetaNODE uses ODEs for meta-optimization on mean prototypes while we use ODEs for generating a representative prototype for each class. 
% \todo{Why we cannot compare with MetaNODE ?}
% More recently, few works focus on designing
% network architecture that does not require retraining the
% model. Squeeze and excite~\cite{roy2020squeeze} proposes a few-shot
% learning architecture specifically designed for medical image
% segmentation. They propose to use squeeze and excite
% modules to fuse information from support image on to query
% image to guide the segmentation arm.~\cite{tang2021recurrent} propose a new framework for
% few-shot medical image segmentation based on prototypical
% networks. 
%They use a context relation encoder and a recurrent mask refinement module to refine the segmentation mask iteratively.\\

\subsection{Adversarial robustness:}
Adversarial attacks for natural image classification have been extensively explored, and interest is turning towards the effects of adversarial attacks in the medical domain~\cite{xu2022medrdf,stimpel2019multi}. FGSM~\cite{goodfellow2014explaining} and PGD~\cite{madry2017towards} generate adversarial examples based on the CNN gradients. Besides image classification, several attack
methods have also been proposed for semantic segmentation~\cite{moosavi2017universal,xie2017adversarial,qi2021stabilized,ozbulak2019impact}.
% object detection and object tracking~\cite{jia}. In~\cite{fischer2017adversarial,Dong2}, the classification based attacks were shown
% transferable to attack deep image segmentation results. 
% The universal perturbations were demonstrated 
% existing in~\cite{moosavi2017universal}.
\cite{xie2017adversarial} introduced Dense Adversary Generation (DAG) that optimizes a loss function over a set of pixels for generating adversarial perturbations.
% method for both semantic segmentation and object detection attacks. 
% They were first
% to make adversarial examples for semantic segmentation.
%\cite{paschali2018generalizability} studied the effects of adversarial attacks on brain segmentation and
%skin lesion classification. 
%They showed that state-of-the-art networks such as~\cite{Inception} and~\cite{Unet} are
%still extremely susceptible to adversarial examples for skin lesion and brain segmentation. 
%The general idea of natural image attacks was to iteratively generate perturbations based on the CNN
%gradients to maximize the network predictions of adversarial examples and the phi labels.
%~\cite{Finlayson} used PGD white and black box attacks on fundoscopy, dermoscopy, and chest X-ray images, using a pre-trained ResNet50 model. %By producing crafted mask, an adaptive segmentation mask attack (ASMA) is proposed to fool DNN model~\cite{ozbulak2019impact}. 
Recently, \cite{qi2021stabilized} proposed an adversarial attack (SMIA) for images in the medical domain that employs a loss stabilization term to exhaustively search the perturbation space.
% to consistently produce adversarial perturbations on medical images.
While adversarial attacks expose the vulnerability of deep neural networks, adversarial training~\cite{madry2017towards,goodfellow2014explaining,liu2020no} is effective in enhancing the target model by training it with adversarial samples.
%~\cite{xu2021dynamic} introduced a standard adversarial training procedure for semantic segmentation.
However, none of the existing methods has explored SAT procedure for few-shot semantic segmentation. In addition to the explicit use of adversarial samples during training, many methods have also explored the use of simple and fast Gaussian perturbations \cite{li2019certified,he2019parametric,ford2019adversarial, byun2022effectiveness,qin2021random} as effective regularisation mechanisms for boosting adversarial robustness. Augmenting the training set with examples perturbed using Gaussian noise to increase adversarial robustness has been suggested in~\cite{zantedeschi2017efficient}.
% ~\cite{fawzi2016robustness} shows that classifiers satisfying curvature constraints are robust to random noise. 
% Later, \cite{ford2019adversarial} shows that  adversarial robustness requires reducing error rates
% to essentially zero under large additive noise.
\cite{li2019certified} uses Renyi
divergence to illustrate the connection of robustness to random noise.~\cite{qin2021random} proposes a defence method against query-based black-box attacks, Random Noise Defense (RND), which adds proper Gaussian noise to each query. RND can be combined with existing defence methods like adversarial training to further boost the adversarial robustness.
% Besides empirically showing better performance with adversarial perturbations, many methods \cite{lecuyer2019certified,li2020sok,ye2020safer} also explicitly provide guarantees of robustness to norm-bounded attacks, leading to certified robustness.~\cite{li2019certified} establishes a connection between robustness against adversarial perturbation and additive random noise, and proposes a training strategy that can significantly improve the certified bounds. ~\cite{li2020sok} provides a comprehensive benchmark on existing robustness verification and training approaches. ~\cite{ye2020safer} describe a structure-free method that  can  be applied to any model, while~\cite{lecuyer2019certified} uses a cryptographically-inspired privacy formalism for  its defense. Again, certified robustness in the data-critical FSS  domain remains largely under-explored.  

Other than external procedures to make models robust, recent studies also propose models that  are intrinsically robust.~\cite{li2020enhancing} uses a Feature Pyramid Decoder with denoising and image restoration that can be applied to any general CNN, improving robustness intrinsically.~\cite{xie2019feature} uses blocks that denoise intermediate features, effectively restoring perturbed features back to the clean ones. The intrinsic adversarial robustness of Neural-ODEs~\cite{chen2018neuralode,kang2021stable,liu2020does,jimenez2022lyanet,huang2021adversarial} is being increasingly explored in recent years, making it an exciting  field of research. 
\begin{figure*}[t!]
\centering
\includegraphics[width=5.05in, height=2.65in]{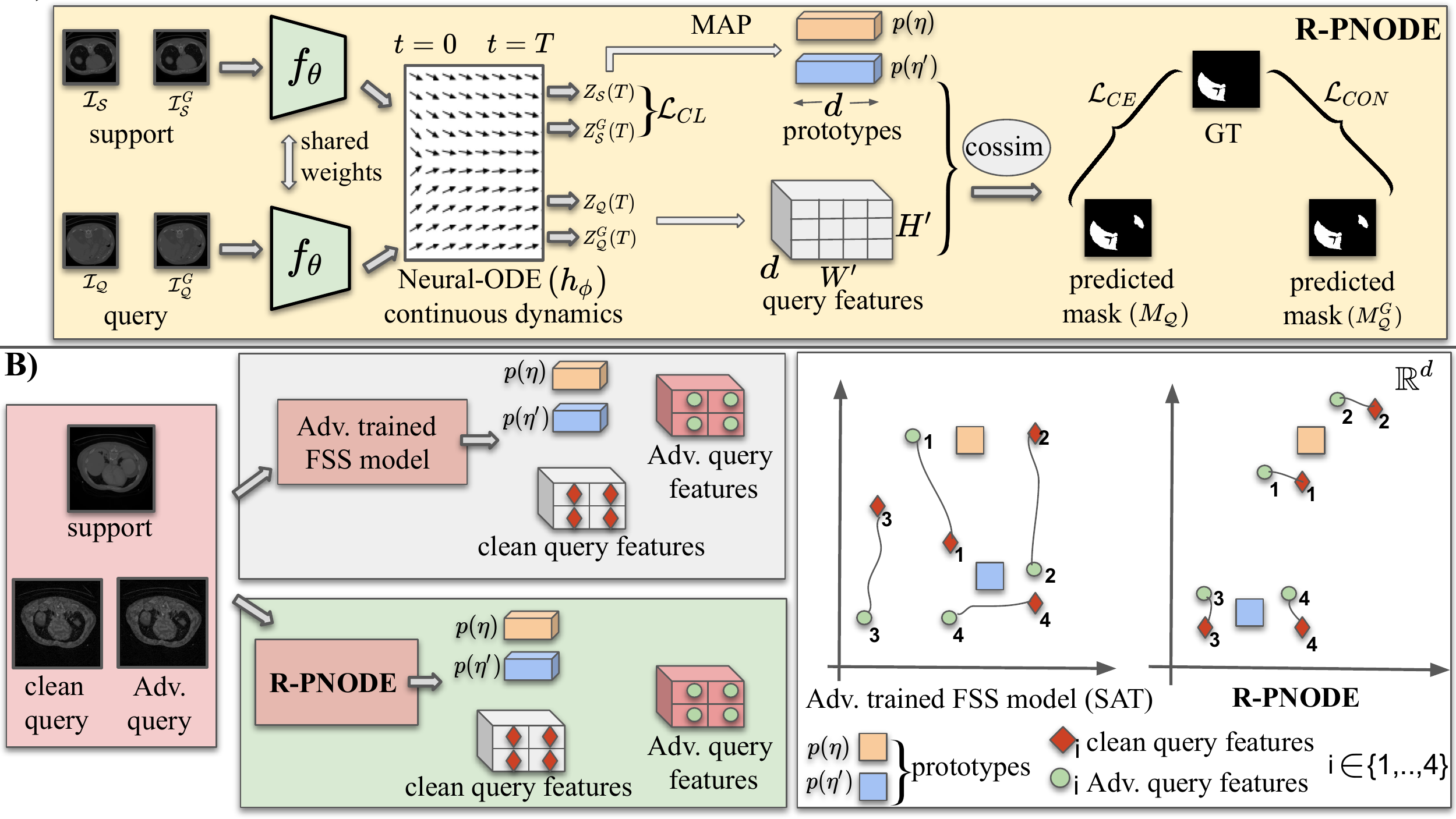}
% \caption{\textbf{A)} The feature extractor $f_\theta$ produces intermediate representations $Z^{k}_{S}$ and query $Z_{Q}$ for support and query images. $t=0$ for the Neural ODE. The final output features $Z_{S}(T)$ and $Z_{Q}(T)$ at time T. These output features depend on the initial states $Z_{S}(0)$,$Z_{Q}(0)$  and the dynamics of the Neural-ODE denoted by $h_{\phi}$ (weights of the neural ODE layers). Prototypes for foreground class $\eta \in \mathrm{C}$ and background class $\eta' $ are obtained by applying Masked Average Pool on the support features $Z_{S}(T)$ and their corresponding label masks. The query mask prediction is done by computing the cosine similarities between all the pixel features in $Z_{Q}(T)$ and the obtained prototypes. \textbf{B)} The working of PNODE is demonstrated by visualizing the d-dimensional representation space to which the clean and adversarial query features are mapped. Due to the non-intersecting property of integral curves in neural ODEs, intrinsic constraints are imposed on the features which make adv. query feature outputs move closer the the clean query feature outputs unlike traditional FSS models. Further, we highlight the limitations of traditional defense methods (Standard Adv. Training) by attacking the adversarially trained models with attacks they aren't trained on to show that PNODE is better at handling various unseen attacks. Similarly, these attacks can be extended to support as well.  
% }
\caption{\textbf{A)} For each support and query image, corresponding gaussian perturbed samples are obtained. Robust features for query and support images (original and gaussian) are obtained by the feature extractor ($f_\theta$) followed by the continuous dynamics and integral solutions of a Neural-ODE. Class-wise prototypes are obtained by applying the Masked Average Pool (MAP) on the clean support features. Pixel-level cosine similarities of query (original and gaussian) features with the prototypes provide corresponding query mask predictions. The support gaussian features are used for cluster loss $\mathcal{L}_{CL}$ (or $\mathcal{L}_{CE}$) while the query gaussian mask predictions are used for consistency loss $\mathcal{L}_{CON}$. \textbf{B)} $d$-dimensional representation of clean and adversarial query features. In R-PNODE, adversarial query features lie closer to the clean ones, unlike SAT. In SAT, perturbations from one class may not be closer to the prototypes of the same class (the model is confused), while  for R-PNODE, they tend to remain close. 
}
\label{fig:pnode_fig}
\end{figure*}
\section{Proposed Method}
The objective is to build a highly accurate few-shot organ segmentation model that is also robust to adversarial attacks. After formalising the problem setting, our methodology focuses primarily on two aspects. First, we describe our proposed method R-PNODE and discuss why it should perform better than existing baselines. Next, we describe and extend SAT for FSS to serve as a strong baseline for comparing adversarial robustness, while also highlighting subtle differences that R-PNODE has from these traditional methods and how this help alleviates some of their drawbacks. 
% First, we propose our framework, R-PNODE, which improves the robustness over SAT while also alleviating some limitations faced by SAT.
%  we extend SAT to the FSS domain as a baseline defence mechanism. Second,
\subsection{Problem setting}
 %The objective is to train a robust few shot segmentation model on the training set $\mathcal{D_{\mathrm{train}}}$ and evaluate it on test set $\mathcal{D_{\mathrm{test}}}$. $\mathcal{D_{\mathrm{train}}}$ and $\mathcal{D_{\mathrm{test}}}$ are constructed using class sets $C_{\mathrm{train}}$ and $C_{\mathrm{test}}$ respectively.
FSS setting includes train $\mathcal{D}_{\mathrm{train}}$ and test $\mathcal{D}_{\mathrm{test}}$ datasets having non-overlapping class sets. 
% ($C_{\mathrm{train}} \cap C_{\mathrm{test}} = \phi$)
% $C_{\mathrm{train}}$ and $\mathcal{C_{\mathrm{test}}}$ are mutually exclusive i.e. $C_{\mathrm{train}} \cap C_{\mathrm{test}} = \phi $.
% The train and test datasets are formed using 
% Both the training $\mathcal{D_{\mathrm{train}}}$ set and the test $\mathcal{D_{\mathrm{test}}}$ set consist
% of several episodes such that each episode comprises of a support set $\mathcal{S}$
% % with its
% % corresponding binary annotation map $L_{S}(\eta)$ with respect to
% % the semantic class (or organ) $\eta$ 
% and a query set $\mathcal{Q}$.
 Each dataset consists of a set of episodes with each episode containing a $N$-way $K$-shot task $\mathcal{T}_{i}$ =  ($\mathcal{S}_{i}, \mathcal{Q}_{i}$) where $\mathcal{S}_i$ and $\mathcal{Q}_i$ is support and query sets for the $i^{th}$ episode having a class set $C_{i}$.
Formally, $\mathcal{D_{\mathrm{train}}} = \{(\mathcal{S}_i,\mathcal{Q}_i)\}^{E_{\mathrm{train}}}_{i = 1}$ and $\mathcal{D_{\mathrm{test}}} = \{(\mathcal{S}_i,\mathcal{Q}_i)\}^{E_{\mathrm{test}}}_{i = 1}$ where $E_{\mathrm{train}}$ and $E_{\mathrm{test}}$ denote the number of episodes during training and testing.
%The semantic classes for training and testing
% are mutually exclusive i.e. $C_{\mathrm{train}} \cap C_{\mathrm{test}} = \phi $.\\
The support set $\mathcal{S}_i$ has $K$ image $(\mathcal{I}_{\mathcal{S}})$, mask $(L_{\mathcal{S}})$ pairs per class
with a total of $N$ semantic classes  i.e. $\mathcal{S}_i =
\{(\mathcal{I}_{\mathcal{S}}^{k},L_{\mathcal{S}}^{k}(\eta))\}$ where  $L^{k}_{\mathcal{S}}(\eta)$ is the ground-truth mask for $k$-th shot corresponding to class$~\eta \in C_{i}$, $|C_{i}| = N$ and $k=1,2, \hdots, K$.
%  Hence for any episode, the support set $\mathcal{S}_i $ contains $N \times K$ samples 
The query set $\mathcal{Q}_i$ has $N_{\mathcal{Q}}$ $\mathcal{\mathrm{image}~(I_{Q})},
{\mathrm{mask}~(L_{\mathcal{Q}})}$ pairs.
%The objective is to learn a semantic segmentation network %$\mathcal{F(.)}$ from
%$\mathcal{D_{\mathrm{train}}}$, such that given a support set %$\mathcal{S}_{\mathrm{novel}} = \{(\mathcal{I}_{S},L_{S}(\eta'))\} \notin \mathcal{D_{\mathrm{train}}}$
%for a novel semantic class $\eta' \in C_{\mathrm{novel}}$ and a query image $\mathcal{I}_Q$,
%we can predict the binary segmentation $M_Q(\eta')$ of the query.\\
%ction{Adversarial Attacks}
% Given an episode $i$ with few-shot semantic segmentation network $\mathcal{F(.)}$, support set ($\mathcal{S}_{i}$) and query image-mask pair ($\mathcal{I}_{Q}$,$L_{Q}$$(\eta)$), the output is a segmentation mask ${L'_{Q}} = \mathcal{F(S, I_Q)}$, where
% $\mathcal{I}_{S}^{k},\mathcal{I_{Q}}\in\mathbb{R}^{H \times W \times 1}$ and ${L_Q} \in \mathbb{R}^{H \times W\times (N+1)}$ - $H,W$ and $N$ are the height,
% width and number of classes in $\mathcal{I}_{Q}$ respectively. \\
The FSS model $\mathcal{F}(.)$ is trained on $\mathcal{D}_{\mathrm{train}}$ across the episodes with support sets and query images as inputs, and predicts the segmentation mask $M_{\mathcal{Q}}$ = $\mathcal{F}(\mathcal{S}_i, \mathcal{I}_{\mathcal{Q}})$ in the $i$-th episode for query image $\mathcal{I}_\mathcal{Q}$. During testing, the trained model $\mathcal{F}(.)$ is used to predict masks for unseen novel classes with the corresponding support set samples and query images as inputs from $\mathcal{D}_{\mathrm{test}}$.
% \todo{added high acc on clean.}with a high accuracy.  
Further, the trained FSS model is adversarially attacked to record the drop in performance. An adversarial version of a clean sample can be generated by exploiting gradient information from the model  $\mathcal{F}(.)$ employing \cite{goodfellow2014explaining}. Specific to the case of FSS, the prediction of query masks not only depends on the query image but also the information from the support set. This enables the attacks to be designed in such a way that either attacked query or support can deteriorate the query prediction. These perturbations are specifically chosen so that the loss between ground truth and the predicted masks of the query increases. 
\subsection{Regularized Prototypical Neural-ODE (R-PNODE)}
The proposed R-PNODE is based on existing prototypical few-shot segmentation models \cite{dong2018few,wang2019panet}. 
Given an episode $i$ with task $\mathcal{T}_{i}$ =  ($\mathcal{S}_{i}, \mathcal{Q}_{i}$),
the feature extractor $f_{\theta}$ generates intermediate feature representations ${Z}^{k}_\mathcal{S}$ and ${Z}_\mathcal{Q}$ for the support and query images $\mathcal{I}^{k}_{\mathcal{S}}$,  $\mathcal{I}_{\mathcal{Q}}$.  
% The Neural-ODE block considers features as a function of time. 
The outputs from the feature extractor $f_{\theta}$ are considered as initial states for the Neural-ODE block at time $t=0$, denoted as ${Z}^{k}_{\mathcal{S}}(0)$ and ${Z}_{\mathcal{Q}}(0)$ respectively. 
% Outputs from the feature extractor $f_{\theta}$ are considered as the initial states for the solution of the Neual-ODE.
From each support and query image, an additional image is generated by applying multiplicative H$\times$W dimensional i.i.d. Gaussian noise, denoted mathematically by:
\begin{equation}
\begin{split}
\mathcal{I}^{kG}_{\mathcal{S}} = \mathcal{I}^{k}_{\mathcal{S}} + \mathcal{I}^{k}_{\mathcal{S}}\odot M \mathrm{\ \ where\ \ } M \sim N_{H\times W}(0,\sigma^2)
\end{split}
\end{equation}
\begin{equation}
\begin{split}
\mathcal{I}^{G}_{\mathcal{Q}} = \mathcal{I}_{\mathcal{Q}} + \mathcal{I}_{\mathcal{Q}}\odot M \mathrm{\ \ where\ \ } M \sim \mathcal{N}_{H\times W}(0,\sigma^2)
\end{split}
\end{equation}
where $\odot$ is the Hadamard operator, for element-wise multiplication. The standard deviation $\sigma$ is a hyperparameter. Similar to the clean images, their corresponding Gaussian samples are also passed through the feature extractor to obtain intermediate features ${Z}^{kG}_{\mathcal{S}}(0)$ and ${Z}^{G}_{\mathcal{Q}}(0)$. 
The Neural-ODE block consists of hidden layers $h_{\phi}$ parameterized by $\phi$ and its dynamics are governed by $h_{\phi}$ which control how the intermediate state changes at any given time $t$. The output representation at fixed terminal time $T (T>0)$ for query features $Z_\mathcal{Q}$ is given by:
%$Z_{\mathcal{Q}}(T) = Z_{\mathcal{Q}}(0) +\int ^{T}_{0}h_{\phi }(Z_{\mathcal{Q}}(t),t)dt$.
\begin{equation} \label{eq:ODE}
\begin{split}
    Z_{\mathcal{Q}}(T) = Z_{\mathcal{Q}}(0) +\int ^{T}_{0}h_{\phi }(Z_{\mathcal{Q}}(t),t)dt  \\  \mathrm{where} \quad  \frac{d{Z_{\mathcal{Q}}}(t)}{d\mathrm{t}} = h_{\phi}({Z}_{\mathcal{Q}}(t), t)
\end{split}
\end{equation}

Similarly, the output representation at fixed terminal time $T (T>0)$ for support features $Z^{k}_{\mathcal{S}}$ are generated.
The support feature maps $Z^{k}_{\mathcal{S}}(T)$ from the Neural-ODE block of spatial dimensions ($H' \times W'$) are upsampled to the same spatial dimensions of their corresponding masks ${L}_{\mathcal{S}}$ of dimension $(H\times W$). 
% In the literature the prototypes are either formed using Global Average Pooling or Masked Average Pooling (MAP). 
Inspired by late fusion~\cite{wang2019panet} where the ground-truth labels are masked over feature maps, we employ Masked Average Pooling (MAP) between
$Z^{k}_{\mathcal{S}}(T)$ and ${L}^{k}_{\mathcal{S}}(\eta)$ to form a $d$-dimensional prototype $p(\eta)$ for each foreground class $\eta \in C_{i}$ as shown: 
% The prototype mean among all the K examples is considered as the final prototype.
\begin{equation}\label{eq:MAP}
    p(\eta)=\dfrac{1}{K}\sum_{k}\dfrac{\sum_{x,y}\{Z^{k}_{\mathcal{S}}(T)\}^{(x,y)}\cdot \mathbbm{1}{[ \{L^{k}_{\mathcal{S}}(\eta)\}^{(x,y)}=\eta]}}{\sum _{x,y}\mathbbm{1}{[ \{L^{k}_{\mathcal{S}}(\eta)\}^{(x,y)}=\eta]}}
\end{equation}
where $(x,y)$ are the spatial locations in the feature map and $\mathbbm{1}(.)$ is an indicator function. 
The background is also treated as a separate class and the prototype for it is calculated by computing the feature mean of all the spatial locations excluding the ones that belong to the foreground classes.
% Given a support set $\mathcal{S}_i = \{(\mathcal{I}_{S}^{k},L^{k}_{S}(\eta))\}$, the feature map output of $h_{\phi}(f_{\theta}())$ for $\mathcal{I}_{S}^{k}$ is $Z^{k}_{\mathcal{S}}$, where $\eta$ $\in C$  and $k=1,2, \hdots, K$.
% In \eqref{eq:MAP}, $(x,y)$ are the spatial locations in the feature map and $\mathbbm{1}$ is an indicator function. 
The probability map over semantic classes $\eta$ is computed by measuring the cosine similarity (\textcolor{black}{cossim}) between each of the spatial locations in $Z_{\mathcal{Q}}(T)$
%\todo{replaces query word in suffix with Q} 
with each prototype $p(\eta)$ as given by:
%\todo{Use sumbol used in eq 7} 
\begin{equation}\label{eq:pred}
{{M}}_{\mathcal{Q}}^{(x,y) }(\eta) = \dfrac{\mathrm{exp}(\mathrm{\textcolor{black}{cossim}}(\{Z_{\mathcal{Q}}(T)\}^{(x,y)},p(\eta)) )}{\sum _{\eta'\in \mathcal{C}_{i}}\mathrm{exp}{(\mathrm{\textcolor{black}{cossim}}( \{Z_{\mathcal{Q}}(T)\}^{(x,y)},p(\eta')) }}
\end{equation}
\textcolor{black}{The denominator involves a summation over all foreground classes $\eta' \in C_{i}$.} The predicted mask $M_{\mathcal{Q}}$ is generated by taking the argmax of $M_{\mathcal{Q}}(\eta)$ across semantic classes. We use Binary Cross Entropy loss $\mathcal{L}_{CE}$ between $M_{\mathcal{Q}}$ and the ground-truth mask for training. 

R-PNODE is further optimized using two additional losses. Eq.~\ref{eq:ODE} and Eq.~\ref{eq:pred} hold for the corresponding features of the gaussian perturbed images as well to get $Z_S^{kG}(T)$ and ${M}^{G}$ for the support and query respectively. The cluster loss ($\mathcal{L}_{CL}$) maximises the cosine similarity between each feature $Z$ extracted from clean support samples and their corresponding Gaussian samples as shown below:
\begin{equation}\label{eq:CL}
\begin{split}
    \mathcal{L}_{CL} = 1 - \frac{1}{K}\sum^{K}_{k = 1}d_{k} ~~, \mathrm{where}~d_{k} = \mathrm{\textcolor{black}{cossim}}(Z_S^k(T), Z_S^{kG}(T))
\end{split}
\end{equation}
\textcolor{black}{
The cluster loss helps to bring the support features and their corresponding gaussian perturbations closer in the representation space which generates robust prototypes that are more representative of the class. This helps to learn better relationships between support and query examples which results in better mask predictions. This technique of augmenting the support images with different gaussian perturbations is not explored by the previous methods. It improves both generalizability and robustness as observed from the results in Table VIII.}
Further, to make the query features robust, we employ consistency loss ($\mathcal{L}_{CON}$) to minimize the difference between Gaussian perturbed query images (${M}^{G}_{\mathcal{Q}}$) and their corresponding ground-truth labels ($L_{\mathcal{Q}}$). For the sake of uniformity, we extend the same loss $\mathcal{L}_{CE}$ as the objective function. 
\begin{equation}\label{eq:CON}
    \mathcal{L}_{CON} = -\dfrac{1}{N}\sum_{x,y}\sum_{\eta'\in \mathcal{C}}\mathbbm{1}[\{L_{\mathcal{Q}}(\eta')\}^{(x,y)}=\eta']\log {M}^{G(x,y)}_{\mathcal{Q}}(\eta')
\end{equation}
\textcolor{black}{
$\mathcal{L}_{CL}$ helps to predict more realistic masks as predicted masks are dependent on the quality of the prototypes and aids $\mathcal{L}_{CON}$ in matching the predicted masks from gaussian perturbed query images with the ground truth masks which further improves robustness and generalizability.}

For the overall procedure, refer to Algorithm~\ref{alg:RNODE} and Fig.~\ref{fig:pnode_fig}.
R-PNODE's superior performance for few-shot segmentation is attributed primarily to two factors. First is the fact that the integral curves learnt by a Neural-ODE are always non-intersecting. We expect a good model to be such that the query and support features of the same class are close to each other. The non-intersecting properties of Neual-ODEs essentially lead to explicit constraints on the features learnt. 
\scalebox{0.75}{
\begin{minipage}{1.3\linewidth}
\centering
\begin{algorithm}[H]
\caption{Regularized Prototypical Neural-ODEs for FSS}
\begin{algorithmic}[1] 
\REQUIRE Clean training data $\mathcal{D_{\mathrm{train}}} = \{(\mathcal{S}_i,\mathcal{Q}_i)\}^{E_{train}}_{i = 1}$, segmentation network $\mathcal{F}(.)$.
\FOR{$\mathrm{i} \in \{1,\hdots, E_{train}\}$}
\STATE Sample episode $\mathcal{E}^{\mathrm{orig}}=\{(\mathcal{S}_{\mathrm{i}},\mathcal{Q}_{\mathrm{i}})\}$ from $\mathcal{D_{\mathrm{train}}}$.
\STATE Get gaussian perturbed  samples $\mathcal{S}_{\mathrm{i}}^{G}$ and $\mathcal{Q}_{\mathrm{i}}^{G} $.
\STATE Get prototypes $p({\eta})$ and features $Z_S^k(T)$  from $\mathcal{S}_{\mathrm{i}} $. Use prototypes to get  predictions $M_{\mathcal{Q}_\mathrm{i}}$ and $M_{\mathcal{Q}_\mathrm{i}}^{G}$. 
\STATE Apply $\mathcal{L}_{CE}$ on mask predictions $M_{\mathcal{Q}_\mathrm{i}}$ and $\mathcal{L}_{CON}$ on $M_{\mathcal{Q}_\mathrm{i}}^{G}$ using Eq.~\ref{eq:CON}.
% \STATE Apply consistency loss with the clean and perturbed queries using equation~\ref{eq:}.
\STATE Extract features $Z_S^{kG}(T)$ for each perturbed support sample and compare with clean $Z_S^k(T)$ to get $\mathcal{L}_{CL}$  by Eq.~\ref{eq:CL}.
\STATE Backpropagate losses $\mathcal{L}_{CE} + \alpha.\mathcal{L}_{CON} + \beta.\mathcal{L}_{CL}$ and update parameters of $\mathcal{F}(.)$.
\ENDFOR
\end{algorithmic}
\label{alg:RNODE}
\end{algorithm}
\end{minipage}%
}
\\

The query features are constrained by the surrounding integral curves of those support prototype features that belong to the same class. Thus, the query features tend to be closer to the prototype features in the representation space. This leads to the accurate classification of query features which ultimately leads to the accurate segmentation of the query images. Fig.~\ref{fig:rpnode_comp} tries to demonstrate this very process. 
% \todo{updated reasoning for clean performance and robustness}

The second factor contributing to the performance of R-PNODE is the addition of cluster and consistency losses. As shown in Fig.~\ref{fig:rpnode_comp}, the gaussian samples serve as additional reference points in the feature space, and by bringing the final features of these closer to the original, the final features of all samples within this larger space are essentially constrained to be very close to each other. 

Extending the reasoning for the superior FSS performance of R-PNODE, the adversarial robustness is again attributed to the intrinsic property of non-intersecting curves of Neural-ODEs along with the proposed losses. As the integral curves of the support features imposed constraints on the features of the query, in a similar way, the clean features also impose constraints on the features of adversarially perturbed images. Thus, features for perturbed support and query tend to lie closer to the features of the clean ones giving rise to adversarial robustness. 
An important difference from the usual adversarial perturbation methods is how the support is perturbed here. Although the ground-truth masks for support images are also available, it makes more sense in the FSS task to take the loss of query prediction using the corresponding support samples. Thus, in Eq.~\ref{eq:FGSM_FSS1}, the loss is between the query mask and query label, but the gradients are calculated w.r.t the support image. We take batches of either the  query or the support perturbed, but not both together. 
%\begin{equation}
%    \mathcal{I^{\mathrm{adv}}_Q} = \mathcal{I_Q} + %\epsilon.\mathrm{sign}(\nabla_{\mathcal{I_Q}}\mathcal{L(F}(\mathcal{I_S}%, \mathcal{I_Q}), L(\eta)))\label{eq:FGSM_FSS2}
%\end{equation}
%On applying this to $\mathcal{D^{\mathrm{orig}}_{\mathrm{train}}}$, we create $\mathcal{D^{\mathcal{Q}}_{\mathrm{train}}}$.\\
% The detailed procedure of SAT is listed in Algorithm~\ref{alg:SAT}.
%In each iteration of SAT, we first calculate the loss $\mathcal{L}$ based on $\mathcal{D^{\mathrm{orig}}_{\mathrm{train}}}$. Using the gradients $\nabla_{\mathcal{I^{\mathrm{orig}}_S}}\mathcal{L}, \nabla_{\mathcal{I^{\mathrm{orig}}_Q}}\mathcal{L}$, we generate $\mathcal{D^{\mathrm{adv}}_{\mathrm{train}}} = \{\mathcal{D^{\mathcal{S}}_{\mathrm{train}}}, \mathcal{D^{\mathcal{Q}}_{\mathrm{train}}}\}$ based on methods (1) and (2).
%$\mathcal{F}$ is further trained on $\mathcal{D^{\mathrm{adv}}_{\mathrm{train}}}$ and the model parameters are updated accordingly.
Generating gaussian perturbations is not computationally expensive when compared to generating adversarial perturbations using gradient-based methods like PGD, SMIA or even FGSM. Thus, R-PNODE has a much smaller computational overhead than SAT. Note that the purpose of the Gaussian samples in R-PNODE is not the same as the adversarial samples in SAT. While SAT uses the perturbations to learn better features on the expected test adversarial distribution, R-PNODE uses gaussian perturbations to populate the training input feature space with more samples thereby having more constrained integral curves. Additionally, note that R-PNODE is attack-agnostic, and thus, its robustness should generalize well to multiple attack strategies. Thus, R-PNODE is able to overcome many of the limitations prevalent in adversarial training methods.
% SAT requires prior knowledge on the type of adversarial attacks to include the corresponding samples during training which is not possible in many practical scenarios. Besides, many gradient based attacks like PGD and SMIA are computationally expensive. Including these during training introduces a significant overhead. Additionally, SAT does not have a mechanism to improve robustness on attacks other than those it had seen during training. These limitations motivate the need for better defense mechanisms and robustness that generalises to different attacks. Our proposed R-PNODE framework described next does not suffer from these shortcomings.
\section{Implementation Details} 
\begin{figure}[!h]
    \centering
    \includegraphics[width=0.99\linewidth]{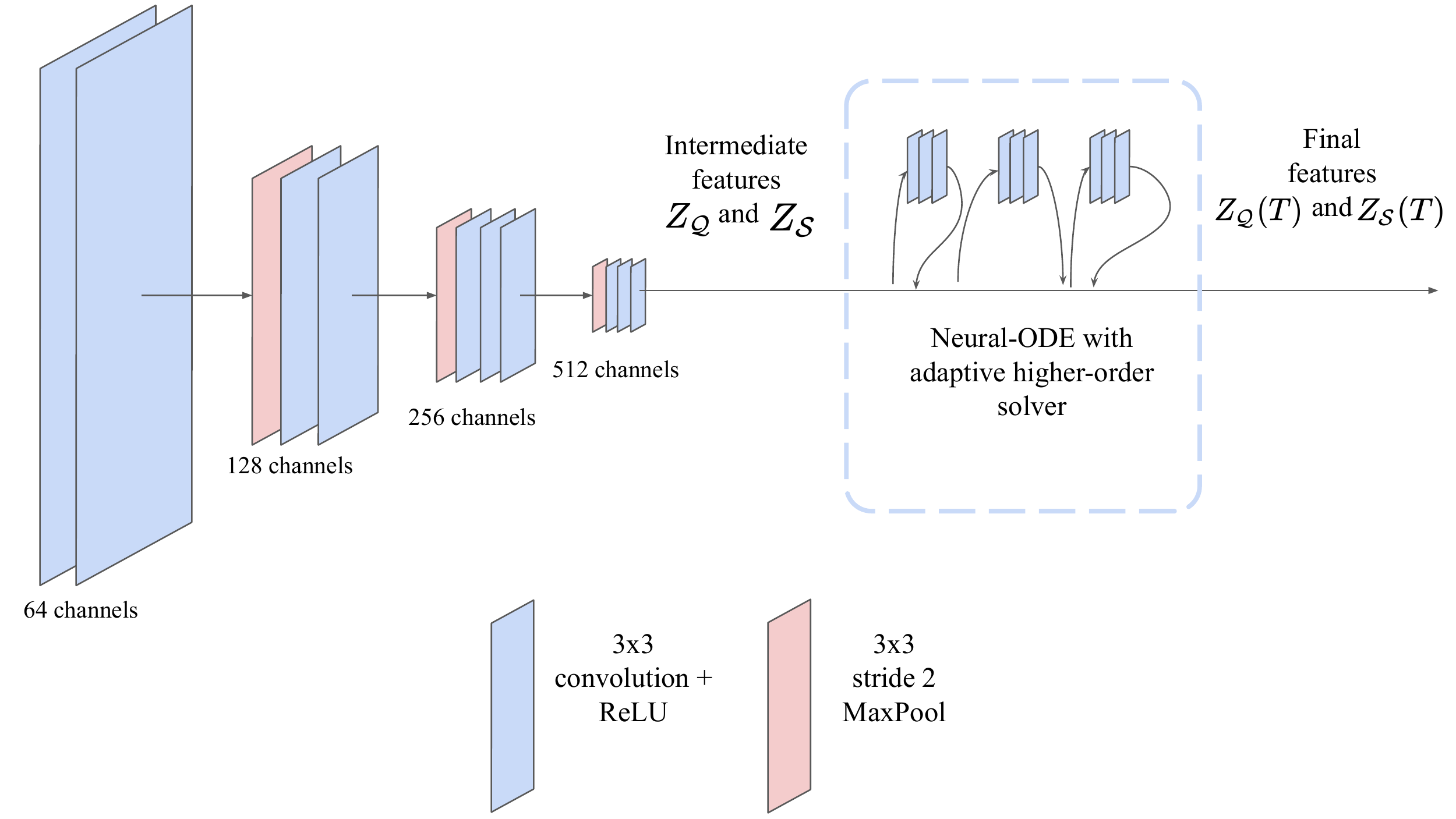}
    \caption{
    \color{black}
    The architecture of the feature extractor $f_{\theta}$ along with Neural-ODE used in R-PNODE. It starts with a VGG-like design and then goes into a Neural-ODE block. The last 3 layers of a standard VGG-16 encoder are removed and replaced with a Neural-ODE block with the hidden dynamics as these 3 layers and a higher order ODE solver that allows adaptive step sizes.
    \color{black}}
    \label{fig:architecture}
\end{figure}

%Our framework consists of a CNN-based feature extractor followed by a Neural ODE block consisting of 3 convolutional layers. The solution for Neural ODE is obtained by using the ODE solver Runge-Kutta of order 5 of Dormand-Prince-Shampine.
% The extracted features are then passed through a continuous Neural ODE solver consisting of 3 convolutional layers as its hidden unit dynamics function.
%The ODE  solver evaluates from $t=0$ to $t=4$, and the value of the solution at $t=4$ is used for further non-parametric prototypical learning. Our implementation is built on top of \emph{torchdiffeq} \cite{torchdiffeq}, an opensource library of differentiable ODE solvers.
The proposed R-PNODE consists of a CNN-based feature extractor followed by a Neural-ODE block consisting of 3 convolutional layers as its hidden dynamics. The architecture of R-PNODE (Fig.~\ref{fig:architecture}) consists of a total of 14.7M trainable parameters, while the baseline models PANet and FSS-1000 have 14.7M and 18.6M parameters, respectively. \textcolor{black}{Taking motivation from~\cite{wang2019panet, kim2021bidirectional}, we use a hidden dimension of d = 512 (output from VGG).} The higher order Runge-Kutta solver~\cite{chen2018neuralode} is used as the black-box approximate solver for the Neural-ODE block. All noises are added at the input level from a $H\times$W dimensional normal distribution with mean 0 and standard deviation $\sigma = 0.1$. Although we perform our experiments on 1-way \{1,3\}-shot setting, the same  approach may be extended to the general n-way k-shot setting as well. 
% The extracted features are then passed through a continuous Neural ODE solver consisting of 3 convolutional layers as its hidden unit dynamics function.
%The ODE  solver evaluates from $t=0$ to $t=4$, and the value of the solution at $t=4$ is used for further non-parametric prototypical learning. Our implementation is built on top of \emph{torchdiffeq} \cite{torchdiffeq}, an opensource library of differentiable ODE solvers.
% \subsubsection{Adversarial Training}
To understand the effect of adversarial training on prototypical networks, we employ SAT with PANet~\cite{wang2019panet} and name it AT-PANet (Adversarially Trained PANet). 
% We experiment with adversarial training on PANet as our baseline is prototypical network.
AT-PANet is trained with FGSM attack of intensity $\epsilon = 0.025$. To test all the trained models, we perturb the support and query images by setting $\epsilon = 0.02, 0.01~\mathrm{and}~0.04$ for FGSM, PGD and SMIA, respectively. For the iterative adversarial attacks SMIA and PGD, we take 10 iterations each. For BIM, CW and DAG, we choose $\epsilon = 0.02$. We take motivation from~\cite{daza2021towards} for the implementation of Auto-Attack and set  0.1 as the dice threshold and 0.01 as the $\epsilon$. These hyperparameters for the attacks were chosen so as to limit the perturbations to human perceptibility. The same set of attack hyperparameters is used for both support and query attacks. 
\textcolor{black}{The loss weighting parameters $\alpha$ and $\beta$ as mentioned in Algorithm~\ref{alg:RNODE} are set to 0.001 ad 0.01 respectively, using grid search. All the baselines are trained using the exact same setting as in R-PNODE. The train, validation and test sets are exactly identical for all the baselines and R-PNODE. The attack's intensities and crop settings are consistent across the baselines. We have used a comparable feature extractor to R-PNODE for all the baselines.}
We use one A100 GPU to conduct our experiments. For statistical significance, we run each experiment twice and report mean and standard deviations. Our implementations, trained models and processed data can be found \href{https://github.com/rpnode-fss/RPNODE_FSS}{here}.
% \footnote{https://github.com/rpnode-fss/RPNODE\_FSS}}.
% \vspace{-2mm}
\section{Experiments and Results}
% \vspace{-1mm}
\subsection{Dataset Description}
\begin{figure}[!h]
    \centering
    \includegraphics[width=0.95\linewidth]{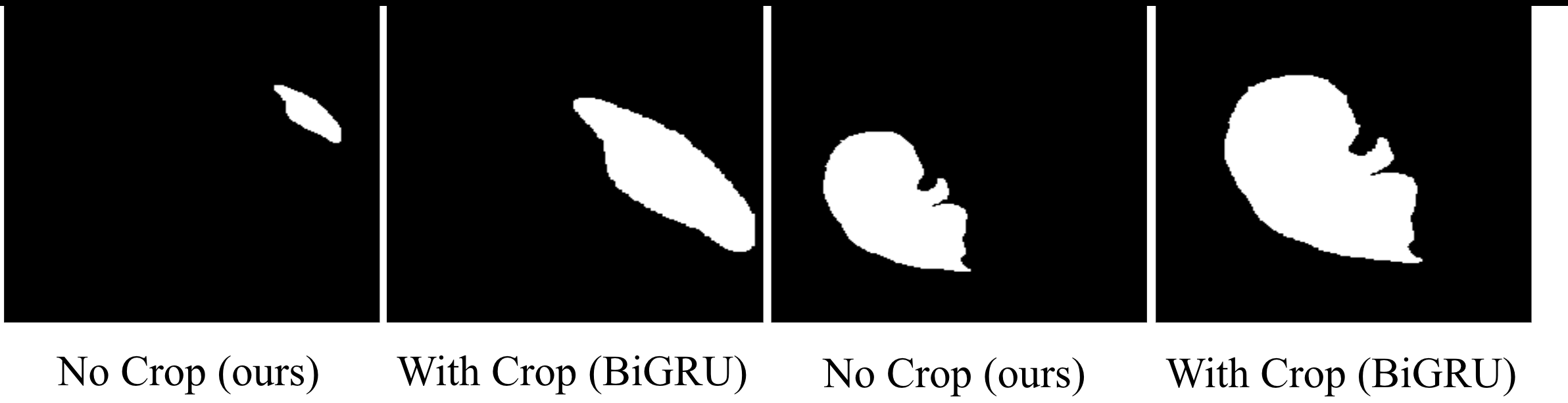}
    \caption{Few samples demonstrating the two datasets crop settings.}
    \label{fig:cropping}
\end{figure}

\textcolor{black}{We experiment on three publicly  available multi-organ segmentation datasets, BCV~\cite{landman2015miccai}, CT-ORG~\cite{rister2020ct}, and Decathlon~\cite{simpson2019large} to evaluate the generalisability of our method. For the in-domain setting, we split BCV into the train (seen) and test (novel) classes.}
% \textcolor{red}
\color{black}
Of the total BCV dataset available, we use half of the dataset for training \textcolor{black}{(6506 slices)}, $\frac{1}{6}\mathrm{th}$ for validation \textcolor{black}{(1779 slices)} and $\frac{1}{3}\mathrm{rd}$ for testing \textcolor{black}{(3873 slices)}. It is to be noted that the dataset splits are done according to the subjects. All slices of any subject will \textcolor{black}{belong to either of the train, validation or test subset. So, the model will not be tested on a slice of any subject it has seen during training.}
\color{black}
For the cross-domain FSS, we train on the BCV dataset (with seen classes) and use novel classes from CT-ORG and Decathlon to test. To have a more uniform size of the test set, we sample 500 random slices per organ from the much larger CT-ORG dataset. \textcolor{black}{We sample 500 and 325 slices from the Liver and Spleen test organs respectively, in the Decathlon dataset.} For the 3D volumes in all three datasets,
% we extract slices having valid masks and divide them to fixed train, test and validation splits. We do not crop the slices class wise and handle multiple organs in the same slice, since cropping in the test set would require labels and this would lead to an unfair evaluation on the test set. 
we extract slices with non-empty masks and divide them into the fixed train, test, and validation splits.~\cite{kim2021bidirectional} cropped the slices based on the ROI for each organ using available masks. We follow a more challenging setting, where no such cropping is applied (Fig.~\ref{fig:cropping}), 
thereby depriving any unfair advantage due to leakage of localisation information. Of the available organs, we report results on the Liver and Spleen (as novel classes) due to their medical significance and availability across datasets. 
\subsection{Choice of Baselines} For baseline comparisons on few-shot organ segmentation, we experiment with traditional FSS methods, PANet~\cite{wang2019panet}, FSS-1000~\cite{li2020fss} and SENet~\cite{roy2020squeeze}. We compare with the recently proposed BiGRU~\cite{kim2021bidirectional}, achieving state-of-the-art results on the selected datasets (in the ROI-cropped setting). Additionally, we compare with a recently proposed Neural-ODE based method SONet~\cite{huang2021adversarial}\footnote{SONet was requiring exorbitant amounts of time for each experiment, so we perform only one run for it.}.
% We also skip one test of SONet with CW attack later for lack of time and resources.}. 
To provide a fair comparison, we use skew-symmetric and input-output stable Neural-ODE blocks of SONet and extend the model to have VGG-like architecture. Finally, to demonstrate adversarial robustness, we compare with AT-PANet, which is PANet~\cite{wang2019panet} trained with the modified adversarial training procedure described in section~\ref{sec:adv_training}.

\subsubsection*{Adversarial Training}\label{sec:adv_training}

\color{black}
To effectively demonstrate the adversarial robustness of R-PNODE, we experiment with Standard Adversarial Training (SAT) as a competitive baseline. \cite{goldblum2020adversarially} recently proposed Adversarial Querying to handle adversarial attacks on the query in few-shot learning. Here, in every task, the query was adversarially perturbed and then standard meta-learning based training was performed with the perturbed query. We first extend this method to segmentation with prototypical FSS models. Next, we modify it to handle both query and support attacks while also focusing on preserving clean accuracy. To accomplish this, we extend $\mathcal{D_{\mathrm{train}}}$ for each batch during training with two additional batches using the update rule from~\cite{goodfellow2014explaining} as follows:  
\begin{enumerate}

\item generate adversarial example  $\mathcal{I^{\mathrm{adv}}_S}$ for support image $\mathcal{I}_{\mathcal{S}}$:
% using the FGSM update rule~\cite{goodfellow2014explaining}:
\begin{equation}
    \mathcal{I^{\mathrm{adv}}_S} = \mathcal{I_S} + \epsilon.\mathrm{sign}(\nabla_{\mathcal{I_S}}\mathcal{L(F}(\mathcal{S}_i, \mathcal{I_Q}), L_{\mathcal{Q}}(\eta)))\label{eq:FGSM_FSS1}
\end{equation}

\item  generate adversarial example $\mathcal{I^{\mathrm{adv}}_Q}$ for query image $\mathcal{I}_{\mathcal{Q}}$:
\begin{equation}
    \mathcal{I^{\mathrm{adv}}_Q} = \mathcal{I_Q} + \epsilon.\mathrm{sign}(\nabla_{\mathcal{I_Q}}\mathcal{L(F}(\mathcal{S}_i, \mathcal{I_Q}), L_{\mathcal{Q}}(\eta)))\label{eq:FGSM_FSS2}
\end{equation}
\end{enumerate}
$\epsilon$ bounds the $l_\infty$ (L-infinity) norm of the perturbation. `sign' is the Signum function. $\nabla_{\mathcal{I}}$ is the gradient of the loss  function $\mathcal{L}$ with respect to the input image $\mathcal{I}$.
While the original batch focuses on training the model for clean samples, the batch from Eq.~\ref{eq:FGSM_FSS1} will improve robustness against support attacks and the batch from Eq.~\ref{eq:FGSM_FSS2} will improve robustness against query attacks. 

\color{black}

\input{clean}
\subsection{Choice of Attacks}
We experiment on 6 different attacks, some traditional and some state-of-the-art. We first use FGSM\cite{goodfellow2014explaining}, one of the most commonly used traditional single-step attack methods. While BIM \cite{kurakin2018adversarial} and PGD\cite{madry2017towards} are some basic iterative methods that extend single-step attacks, CW \cite{carlini2017towards} is a more sophisticated one. Auto-Attack\cite{croce2020reliable} is an  adaptive method involving an ensemble of four diverse attacks to reliably evaluate robustness. While these attacks were originally proposed for classification, we extend them to segmentation in our experiments. DAG \cite{xie2017adversarial} was one of the first methods for adversarial attacks specific to segmentation. SMIA\cite{qi2021stabilized} was another segmentation attack proposed quite recently, that focuses on the medical domain, and is shown to perform better than many other methods. With this broad spectrum of attacks covering the most important categories, we are able to effectively demonstrate the efficacy of our proposed approach. 

\subsection{Results and Discussion}
\subsubsection{Few-Shot Organ Segmentation}
We report the results for 1-shot organ segmentation in Table~\ref{tab:clean} and those for 3-shot in Fig.~\ref{fig:3shot_clean}. R-PNODE outperforms all compared baselines by significant margins across all organs both for in-domain and cross-domain settings. On average, R-PNODE outperforms all baselines by at least 29\%, 20\% and 33\% for BCV in-domain, BCV $\rightarrow$ CT-ORG cross-domain and BCV $\rightarrow$ Decathlon cross-domain, respectively. R-PNODE outperforms the baselines in the 3-shot setting as well, as evident from Fig.~\ref{fig:3shot_clean}. It is able to outperform PANet, AT-PANet and SONet, even though they have an almost identical number of parameters. R-PNODE learns a better representation space of support and query samples owing to the continuous dynamics of the Neural-ODE, resulting in superior performance. 

%#crf results not added.
\textcolor{black}{
Since the Liver organ had a significant number of annotations in the BCV dataset, we tried an experiment training Liver in a fully supervised manner. With the same backbone as R-PNODE and a U-Net style architecture, we observed that even with full supervision, the dice score was 0.84, which is not too high compared to our FSS result of 0.79 (in-domain setting in Table~\ref{tab:clean}) considering the amount of supervision used. Given that we want to automate the segmentation process, this tradeoff is justifiable. Hence, there is a significant scope of methods like ours in the wide spectrum between fully automated and fully manual.
We also compare with feature-based methods like Level Set \cite{JIANG2012840} and CRF \cite{lafferty2001conditional} as shown in Table~\ref{tab:clean}. It can be seen that although Level Set \cite{JIANG2012840} performs well for the Liver organ, both perform quite poorly for the smaller and harder-to-segment Spleen organ.
}

\subsubsection{Adversarial Robustness}
As can be seen in Table~\ref{tab:bcv_1shot_liver} and Table~\ref{tab:bcv_1shot_spleen}, R-PNODE outperforms all of our baselines by at least 15\%, 25\%, 17\%, 16\%, 16\%, 45\% and 10\% on FGSM, PGD, SMIA, BIM, CW, DAG and Auto-Attack query attacks respectively for the BCV in-domain Spleen setting. We report results on the cross-domain setting in Table~\ref{tab:ctorg} and Table~\ref{tab:decathlon}. As is evident from these tables, similar to distribution shifts between clean and perturbed samples, R-PNODE  is also robust to cross-domain distribution shifts. 
% \begin{table}[!ht]
%     \centering
%     \setlength{\tabcolsep}{10pt}
%     \renewcommand{\arraystretch}{1.2}
%     \caption{Query attack results for BCV $\rightarrow$ CT-ORG cross-domain Liver. The dice scores are rounded off to 2 decimals.}
%     \scalebox{1}{
%     \begin{tabular}{ccccc}
%     \toprule
%         Method & Clean & FGSM & PGD & SMIA \\
%     \midrule
%         PANet\cite{wang2019panet} & 0.52 ± 0.01 & 0.16 ± 0.03 & 0.14 ± 0.01 & 0.20 ± 0.01   \\
%         FSS1000\cite{li2020fss} & 0.29 ± 0.04 & 0.05 ± 0.01 & 0.01 ± 0.01 & 0.14 ± 0.02\\
%         SENet\cite{roy2020squeeze}  & 0.47 ± 0.01 & 0.14 ± 0.04 & 0.11 ± 0.02 & 0.15 ± 0.05\\
%       %\midrule
%         AT-PANet & 0.56 ± 0.01 & 0.23 ± 0.08 & 0.14 ± 0.01 & 0.31 ± 0.01   \\
%         \textbf{R-PNODE} & \textbf{0.67 ± 0.01} & \textbf{0.32 ± 0.02} & \textbf{0.22 ± 0.02} & \textbf{0.42 ± 0.02}\\
%     \bottomrule
    
%     \end{tabular}}
%     \label{tab:ctorg}
% \end{table}
% \input{bcv_1shot}
\begin{table*}[!ht]
    \centering
    \setlength{\tabcolsep}{10pt}
    \renewcommand{\arraystretch}{1.2}
    \caption{Query attack results \textcolor{black}{(higher is better)} for BCV $\rightarrow$ BCV \textcolor{black}{1-shot} in-domain for Liver organ (novel class). \\The dice scores are rounded off to two decimals. }

    \scalebox{0.81}{
    \begin{tabular}{cccccccc}
    \toprule
        Method & FGSM & PGD & SMIA & BIM & CW & DAG & Auto-Attack\\
    \midrule
        PANet\cite{wang2019panet} & 0.29 ± 0.01 & 0.21 ± 0.01& 0.20 ± 0.01 & 0.20 ± 0.01 & 0.29 ± 0.04 & 0.19 ± 0.03 & 0.08 ± 0.01\\
        FSS-1000\cite{li2020fss} & 0.10 ± 0.03 & 0.04 ± 0.02& 0.18 ± 0.01 & 0.05 ± 0.04 &  0.33 ± 0.06 & 0.06 ± 0.04 & 0.01 ± 0.01 \\
        SENet\cite{roy2020squeeze} & 0.30 ± 0.06 & 0.22 ± 0.02& 0.12 ± 0.02& 0.19 ± 0.02 & 0.33 ± 0.01 & 0.32 ± 0.01  & 0.07 ± 0.01\\
        BiGRU\cite{kim2021bidirectional} & 0.16 ± 0.12 & 0.05 ± 0.05 & 0.35 ± 0.01 & 0.16 ± 0.01 & 0.08 ± 0.02 & 0.01 ± 0.01  & 0.02 ± 0.02   \\
        SONet\cite{huang2021adversarial} & 0.19 ± 0.00 & 0.11 ± 0.00 & 0.06 ± 0.00 & 0.09 ± 0.00 & 0.00 ± 0.00 & 0.06 ± 0.00 & 0.03 ± 0.01\\
        AT-PANet & 0.35 ± 0.03 & \textbf{0.27 ± 0.02}& 0.36 ± 0.01  &  0.28 ± 0.04 & 0.11 ± 0.06 & 0.31 ± 0.06 & 0.08 ± 0.02\\
        \textbf{R-PNODE} & \textbf{0.41 ± 0.02} & \textbf{0.27 ± 0.01}& \textbf{0.53 ± 0.03} & \textbf{0.29 ± 0.06} & \textbf{0.55 ± 0.07} & \textbf{0.34 ± 0.04} & \textbf{0.10 ± 0.03}\\
    \bottomrule
    
    \end{tabular}}

    \label{tab:bcv_1shot_liver}
\end{table*}

\begin{table*}[!ht]
    \centering
    \setlength{\tabcolsep}{10pt}
    \renewcommand{\arraystretch}{1.2}
    \caption{Query attack results \textcolor{black}{(higher is better)} for BCV $\rightarrow$ BCV \textcolor{black}{1-shot} in-domain for Spleen organ (novel class). \\The dice scores are rounded off to two decimals. }

    \scalebox{0.81}{
    \begin{tabular}{cccccccc}
    \toprule
        Method  & FGSM & PGD & SMIA & BIM & CW & DAG & Auto-Attack \\
    \midrule
        PANet\cite{wang2019panet} &  0.16 ± 0.01 & 0.11 ± 0.01 & 0.07 ± 0.01  &0.12 ± 0.01 & 0.03 ± 0.01 & 0.02 ± 0.01 & 0.06 ± 0.01\\
        FSS-1000\cite{li2020fss} & 0.19 ± 0.01 & 0.08 ± 0.01 & 0.17 ± 0.01  &0.05 ± 0.02 & 0.19 ± 0.10& 0.04 ± 0.04  & 0.04 ± 0.01 \\
        SENet\cite{roy2020squeeze} & 0.04 ± 0.01 & 0.21 ± 0.04& 0.01 ± 0.01 & 0.25 ± 0.05 & 0.25 ± 0.03 & 0.12 ± 0.04  & 0.01 ± 0.01 \\
        BiGRU\cite{kim2021bidirectional} & 0.19 ± 0.01 & 0.03 ± 0.01 & 0.15 ± 0.01 & 0.19 ± 0.01 & 0.01 ± 0.01 & 0.01 ± 0.01 & 0.01 ± 0.01 \\
        SONet\cite{huang2021adversarial} & 0.21 ± 0.00 & 0.08 ± 0.00 & 0.14 ± 0.00 & 0.09 ± 0.00 & 0.00 ± 0.00 & 0.01 ± 0.00 & 0.05 ± 0.01\\
        AT-PANet  & 0.32 ± 0.08 & 0.19 ± 0.03 & 0.11 ± 0.01  &0.16 ± 0.02& 0.02 ± 0.02 &0.06 ± 0.04 & 0.10 ± 0.07\\
        \textbf{R-PNODE} & \textbf{0.37 ± 0.02} & \textbf{0.28 ± 0.03} & \textbf{0.20 ± 0.02} &\textbf{ 0.29 ± 0.05} & \textbf{0.29 ± 0.02} & \textbf{0.22 ± 0.01} & \textbf{0.11 ± 0.02}\\
    \bottomrule
    
    \end{tabular}}

    \label{tab:bcv_1shot_spleen}
\end{table*}

\begin{table}[!ht]
    \centering
    \setlength{\tabcolsep}{10pt}
    \renewcommand{\arraystretch}{1.2}
    \caption{Query attack results \textcolor{black}{(higher is better)} for BCV $\rightarrow$ CT-ORG \textcolor{black}{1-shot} for Liver. The dice scores are rounded off to two decimals.}
    \scalebox{0.8}{
    \begin{tabular}{cccc}
    \toprule
        Method & FGSM & PGD & SMIA \\
    \midrule
        PANet\cite{wang2019panet} & 0.16 ± 0.03 & 0.14 ± 0.01 & 0.20 ± 0.01   \\
        FSS-1000\cite{li2020fss} & 0.05 ± 0.01 & 0.01 ± 0.01 & 0.14 ± 0.02\\
        SENet\cite{roy2020squeeze}  & 0.14 ± 0.04 & 0.11 ± 0.02 & 0.15 ± 0.05\\
      %\midrule
        AT-PANet &  0.23 ± 0.08 & 0.14 ± 0.01 & 0.31 ± 0.01   \\
        \textbf{R-PNODE}  & \textbf{0.32 ± 0.02} & \textbf{0.22 ± 0.02} & \textbf{0.42 ± 0.02}\\
    \bottomrule
    
    \end{tabular}}
    \label{tab:ctorg}
\end{table}
% \begin{table}[!ht]
%     \centering
%     \setlength{\tabcolsep}{2.8pt}
%     \renewcommand{\arraystretch}{1.2}
%     \caption{Query attack results for BCV $\rightarrow$ Decathlon cross-domain Liver. Dice scores are rounded off to 2 decimals.}
%     \scalebox{1}{
%     \begin{tabular}{ccccc}
%     \toprule
%         Method & Clean & FGSM & PGD & SMIA \\
%     \midrule
%         PANet\cite{wang2019panet}  & 0.53 ± 0.01 & 0.18 ± 0.03 & 0.13 ± 0.01 & 0.22 ± 0.01   \\
%         FSS1000\cite{li2020fss} & 0.37 ± 0.06 & 0.09 ± 0.05 & 0.03 ± 0.02 & 0.12 ± 0.02\\
%         SENet\cite{roy2020squeeze}  & 0.50 ± 0.01 & 0.18 ± 0.06 & 0.13 ± 0.02 & 0.14 ± 0.07\\
%       %\midrule
%         AT-PANet & 0.57 ± 0.01 & 0.35 ± 0.04 & 0.19 ± 0.04 & 0.29 ± 0.01\\
%         \textbf{R-PNODE} & \textbf{0.79 ± 0.01} & \textbf{0.40 ± 0.01} & \textbf{0.25 ± 0.04} & \textbf{0.45 ± 0.01}\\
%     \bottomrule
    
\begin{figure*}
    \centering
    \includegraphics[width=0.329\linewidth]{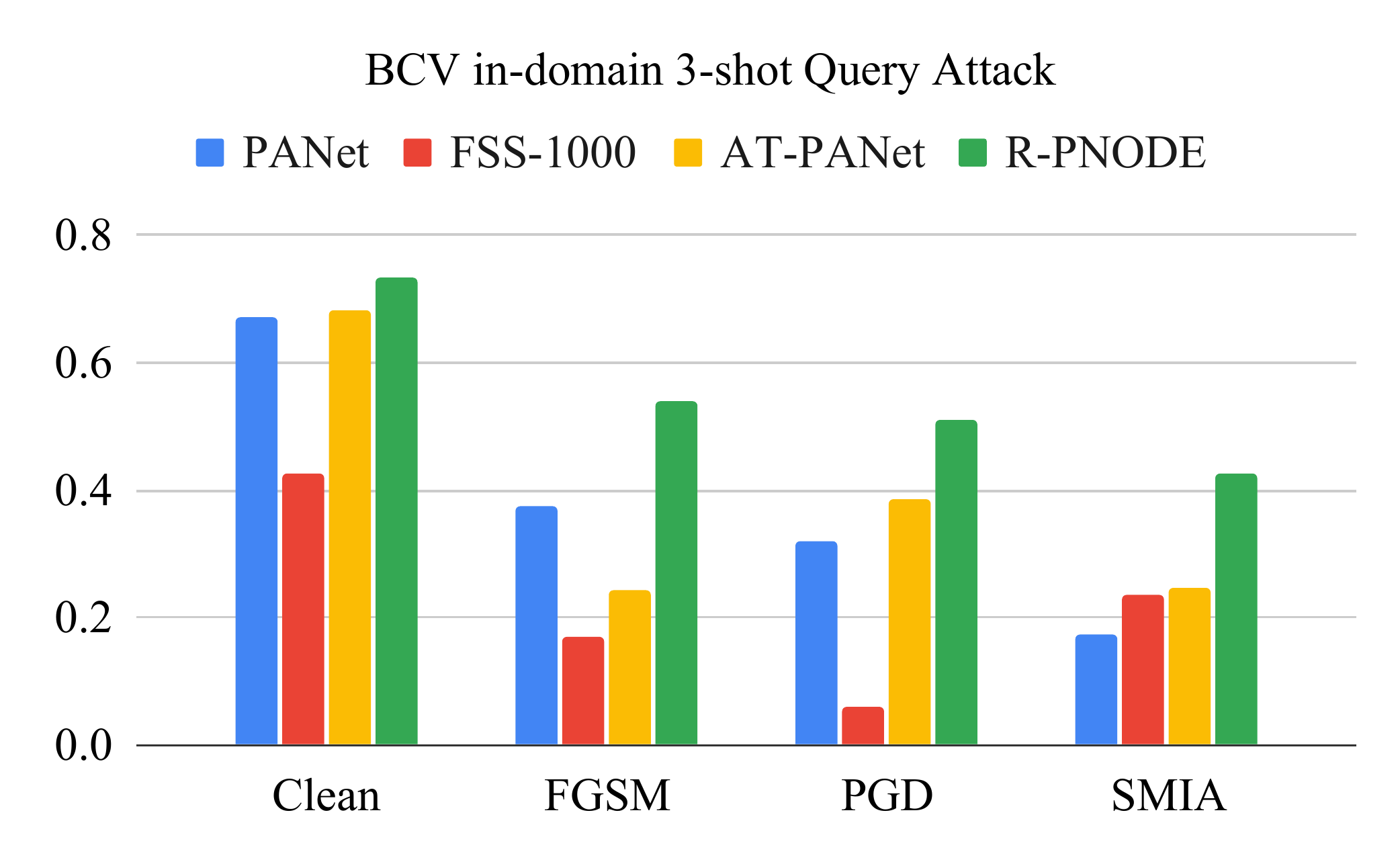}
    \includegraphics[width=0.329\linewidth]{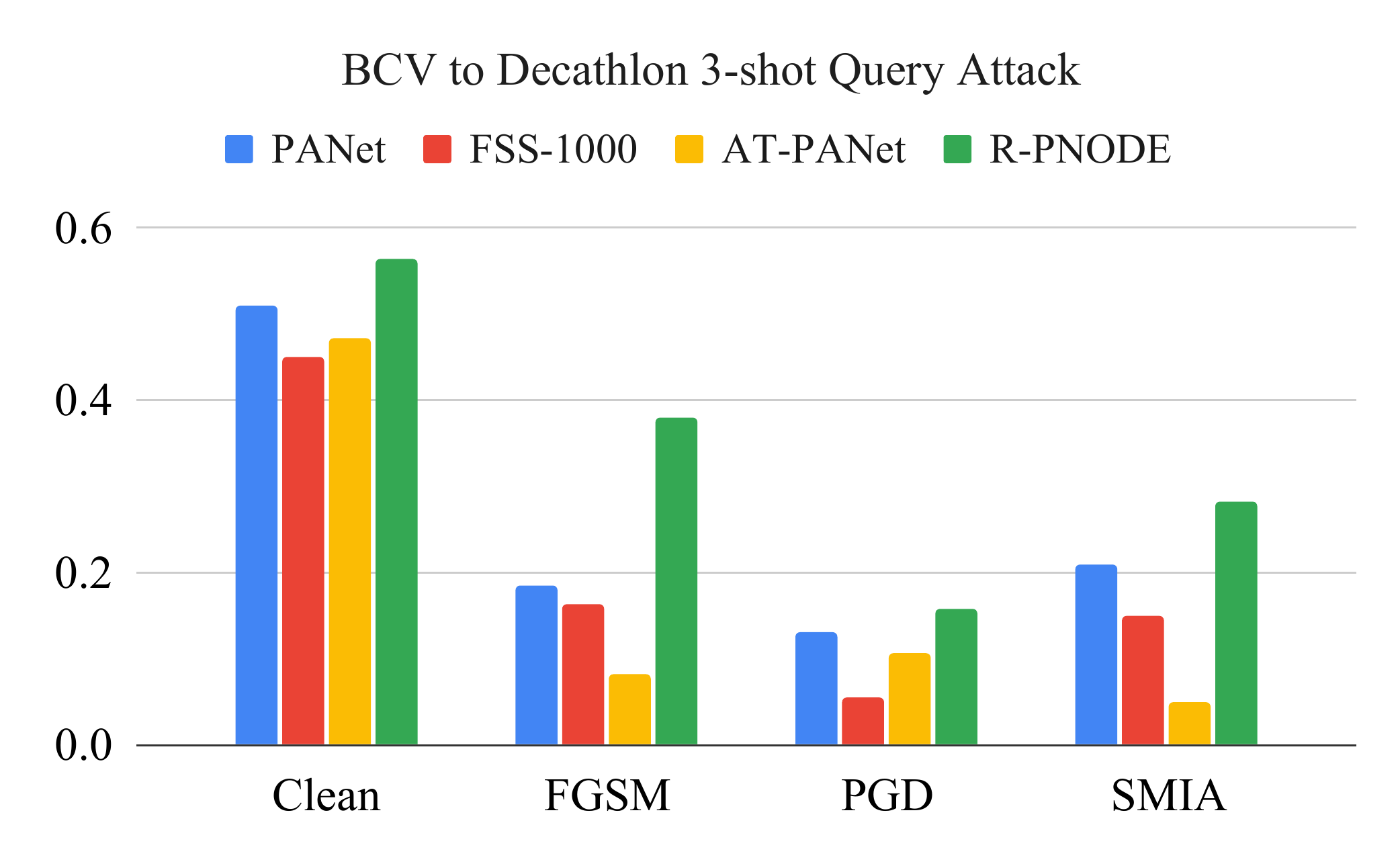}
    \includegraphics[width=0.329\linewidth]{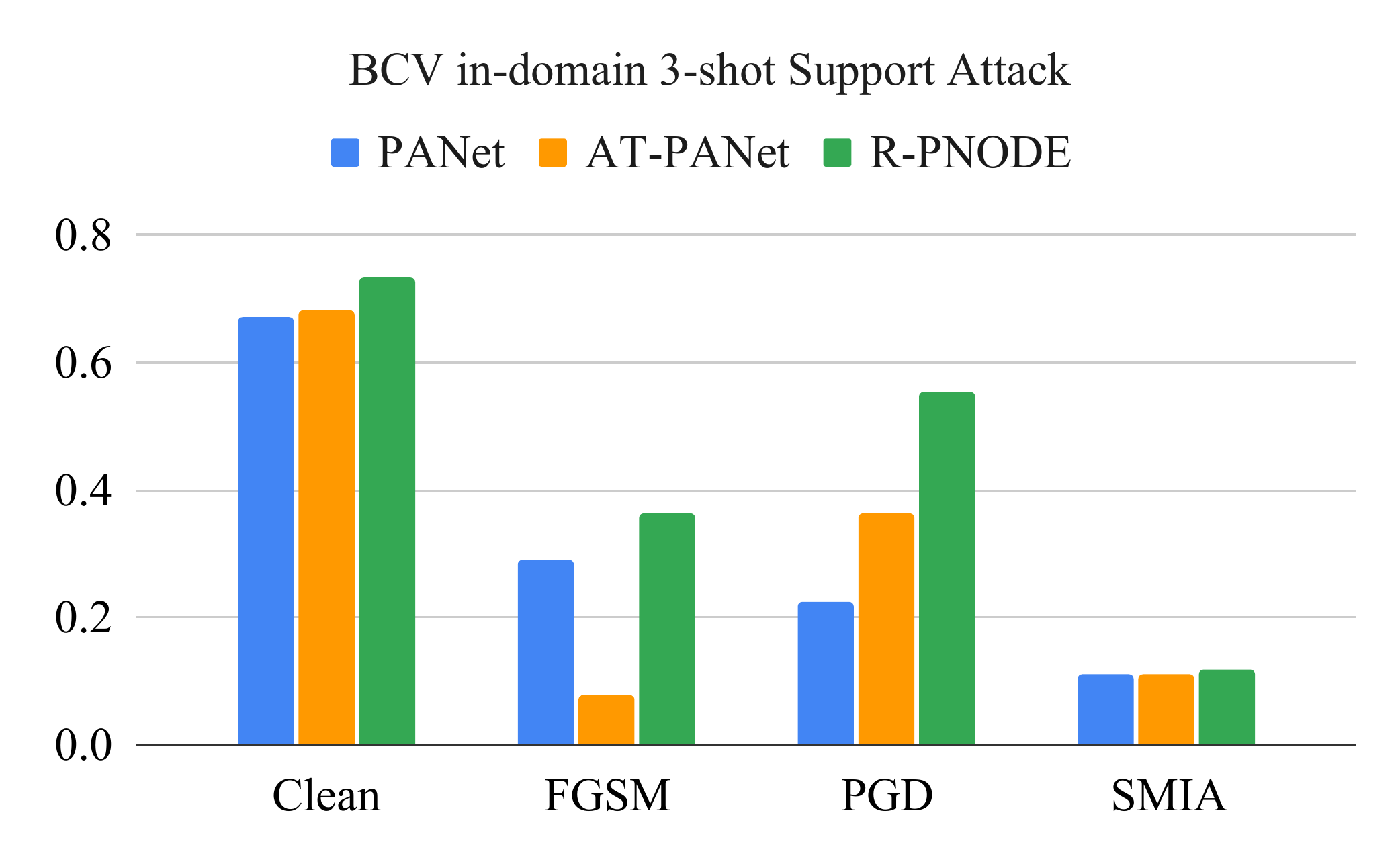}
    \caption{3-shot FSS results for BCV $\rightarrow$ BCV (in-domain) and BCV $\rightarrow$ Decathlon for Liver organ (novel class) when support and query are attacked.}
    \label{fig:3shot}
\end{figure*}

%     \end{tabular}}
%     \label{tab:decathlon}
% \end{table}
\begin{table}[!ht]
    \centering
    \setlength{\tabcolsep}{10pt}
    \renewcommand{\arraystretch}{1.2}
    \caption{Query attack results \textcolor{black}{(higher is better)} for BCV $\rightarrow$ Decathlon \textcolor{black}{1-shot} for Liver. Dice scores are rounded off to two decimals.}
    \scalebox{0.8}{
    \begin{tabular}{cccc}
    \toprule
        Method & FGSM & PGD & SMIA \\
    \midrule
        PANet\cite{wang2019panet}  & 0.18 ± 0.03 & 0.13 ± 0.01 & 0.22 ± 0.01   \\
        FSS-1000\cite{li2020fss} & 0.09 ± 0.05 & 0.03 ± 0.02 & 0.12 ± 0.02\\
        SENet\cite{roy2020squeeze}  & 0.18 ± 0.06 & 0.13 ± 0.02 & 0.14 ± 0.07\\
      %\midrule
        AT-PANet  & 0.35 ± 0.04 & 0.19 ± 0.04 & 0.29 ± 0.01\\
        \textbf{R-PNODE} & \textbf{0.40 ± 0.01} & \textbf{0.25 ± 0.04} & \textbf{0.45 ± 0.01}\\
    \bottomrule
    
    \end{tabular}}
    \label{tab:decathlon}
\end{table}
We also evaluate R-PNODE on support attacks and compare its performance with other prototypical networks. The results of these experiments are reported in Table~\ref{tab:sup1} and Table~\ref{tab:sup2}. We observed much lower deterioration of performance with \textit{non-prototypical} networks with the support attacks and believe that the support attack strategy used by us is not suitable for them. R-PNODE outperforms other prototypical baselines in most settings and improves by a considerable margin in many. Thus, it is adversarially robust to perturbations in the support images too. The improvements for different attacks are much more skewed for support attacks when compared to query attacks. For example, while Liver BCV in-domain improves by over 100\% for PGD attacks, it shows only 9\% improvement on SMIA. In other settings as well, the performance improvement on SMIA is considerably small. The performance on other attacks is so good that they even outperform the clean performance of some baselines. For example, the dice scores on Liver in-domain setting of R-PNODE for FGSM and PGD (Table \ref{tab:sup1}) are higher than that for PANet and AT-PANet for clean data in the same setting (Table \ref{tab:clean}). Another interesting observation is that while AT-PANet outperforms  PANet for all the in-domain experiments, this trend is not followed for the cross-domain ones. This hints towards possible drawbacks of the SAT defence mechanism. In addition to these experiments for the 1-shot setting, we also test the performance in the 3-shot setting. The results in Fig.~\ref{fig:3shot} demonstrate  that R-PNODE performs better than our baselines in this setting as well. 
\begin{table}[!t]
    \centering
    \setlength{\tabcolsep}{6pt}
    \renewcommand{\arraystretch}{1.2}
    \caption{Support attack results \textcolor{black}{(higher is better)} with prototypical networks for BCV $\rightarrow$ BCV \textcolor{black}{1-shot} in-domain for Liver and Spleen as novel classes.}
    \scalebox{0.79}{
    \begin{tabular}{cccccc}
    \toprule
       Organ & Method & FGSM & PGD & SMIA \\
    \midrule
        \multirow{3}{*}{Liver} & PANet\cite{wang2019panet} & 0.17 ± 0.06 & 0.39 ± 0.01 & 0.11 ± 0.01   \\
        & AT-PANet  & 0.19 ± 0.07 & 0.30 ± 0.15 & 0.11 ± 0.01\\
        & \textbf{R-PNODE} & \textbf{0.67 ± 0.01} & \textbf{0.68 ± 0.08} & \textbf{0.12 ± 0.01}\\
    \midrule
        \multirow{3}{*}{Spleen} & PANet\cite{wang2019panet} & 0.17 ± 0.02 & 0.10 ± 0.02 & 0.03 ± 0.01   \\
        & AT-PANet & \textbf{0.42 ± 0.03} & 0.32 ± 0.06 & 0.03 ± 0.01   \\
        & \textbf{R-PNODE} & 0.38 ± 0.01 & \textbf{0.40 ± 0.02} & \textbf{0.04 ± 0.02}\\
    \bottomrule
    
    \end{tabular}}
    \label{tab:sup1}
\end{table}
% \begin{table}[!t]
%     \centering
%     \setlength{\tabcolsep}{1.5pt}
%     \renewcommand{\arraystretch}{1.2}
%     \caption{Support attack results for prototypical networks on cross-domain BCV $\rightarrow$ CT-ORG and BCV $\rightarrow$ Decathlon Liver.}
%     \scalebox{1}{
%     \begin{tabular}{cccccc}
%     \toprule
%         & Method & Clean & FGSM & PGD & SMIA \\
%     \midrule
%         \multirow{3}{*}{\rotatebox[origin=c]{90}{CT-ORG}} & PANet\cite{wang2019panet} & 0.52 ± 0.01 & 0.20 ± 0.05 & 0.43 ± 0.02 & 0.14 ± 0.01   \\
%         & AT-PANet  & 0.56 ± 0.01 & 0.12 ± 0.02 & 0.32 ± 0.08 & 0.14 ± 0.01\\
%         & \textbf{R-PNODE} & \textbf{0.67 ± 0.01} & \textbf{0.51 ± 0.01} & \textbf{0.52 ± 0.02} & \textbf{0.15 ± 0.01}\\
%     \midrule
%         \multirow{3}{*}{\rotatebox[origin=c]{90}{Decathlon}} & PANet\cite{wang2019panet} & 0.53 ± 0.01 & 0.15 ± 0.01 & 0.35 ± 0.02 & 0.13 ± 0.01   \\
%         & AT-PANet & 0.57 ± 0.01 & 0.14 ± 0.04 & 0.31 ± 0.06 & \textbf{0.13 ± 0.01}   \\
%         & \textbf{R-PNODE} & \textbf{0.79 ± 0.01} & \textbf{0.66 ± 0.01} & \textbf{0.66 ± 0.02} & \textbf{0.13 ± 0.01}\\
%     \bottomrule
    
%     \end{tabular}}
%     \label{tab:sup2}
% \end{table}
\begin{table}[!t]
    \centering
    \setlength{\tabcolsep}{6pt}
    \renewcommand{\arraystretch}{1.2}
    \caption{Support attack results \textcolor{black}{(higher is better)} with prototypical networks for BCV $\rightarrow$ CT-ORG and BCV $\rightarrow$ Decathlon \textcolor{black}{1-shot} settings for Liver.}
    \scalebox{0.79}{
    \begin{tabular}{ccccc}
    \toprule
       Dataset & Method & FGSM & PGD & SMIA \\
    \midrule
        \multirow{3}{*}{CT-ORG} & PANet\cite{wang2019panet} & 0.20 ± 0.05 & 0.43 ± 0.02 & 0.14 ± 0.01   \\
        & AT-PANet  & 0.12 ± 0.02 & 0.32 ± 0.08 & 0.14 ± 0.01\\
        & \textbf{R-PNODE} & \textbf{0.51 ± 0.01} & \textbf{0.52 ± 0.02} & \textbf{0.15 ± 0.01}\\
    \midrule
        \multirow{3}{*}{Decathlon} & PANet\cite{wang2019panet} & 0.15 ± 0.01 & 0.35 ± 0.02 & 0.13 ± 0.01   \\
        & AT-PANet  & 0.14 ± 0.04 & 0.31 ± 0.06 & \textbf{0.13 ± 0.01}   \\
        & \textbf{R-PNODE}  & \textbf{0.66 ± 0.01} & \textbf{0.66 ± 0.02} & \textbf{0.13 ± 0.01}\\
    \bottomrule
    
    \end{tabular}}
    \label{tab:sup2}
\end{table}

% 
% \begin{figure}
%     \centering
%     \includegraphics[width=0.49\linewidth]{CTORG 1-shot (Liver).pdf}
%     \includegraphics[width=0.49\linewidth]{Decathlon 1-shot (Liver).pdf}
%     \caption{Performance of models for attacks on BCV $\rightarrow$ CT-ORG (left), BCV $\rightarrow$ Decathlon (middle) and for different intensities of FGSM (right), on Liver 1-shot.}
%     \label{fig:plots}
% \end{figure}
\begin{figure}[!h]
    \centering
    \fboxsep=0.5mm
    \fboxrule=1pt
    \fcolorbox{black}{white}{\includegraphics[width=0.45\linewidth]{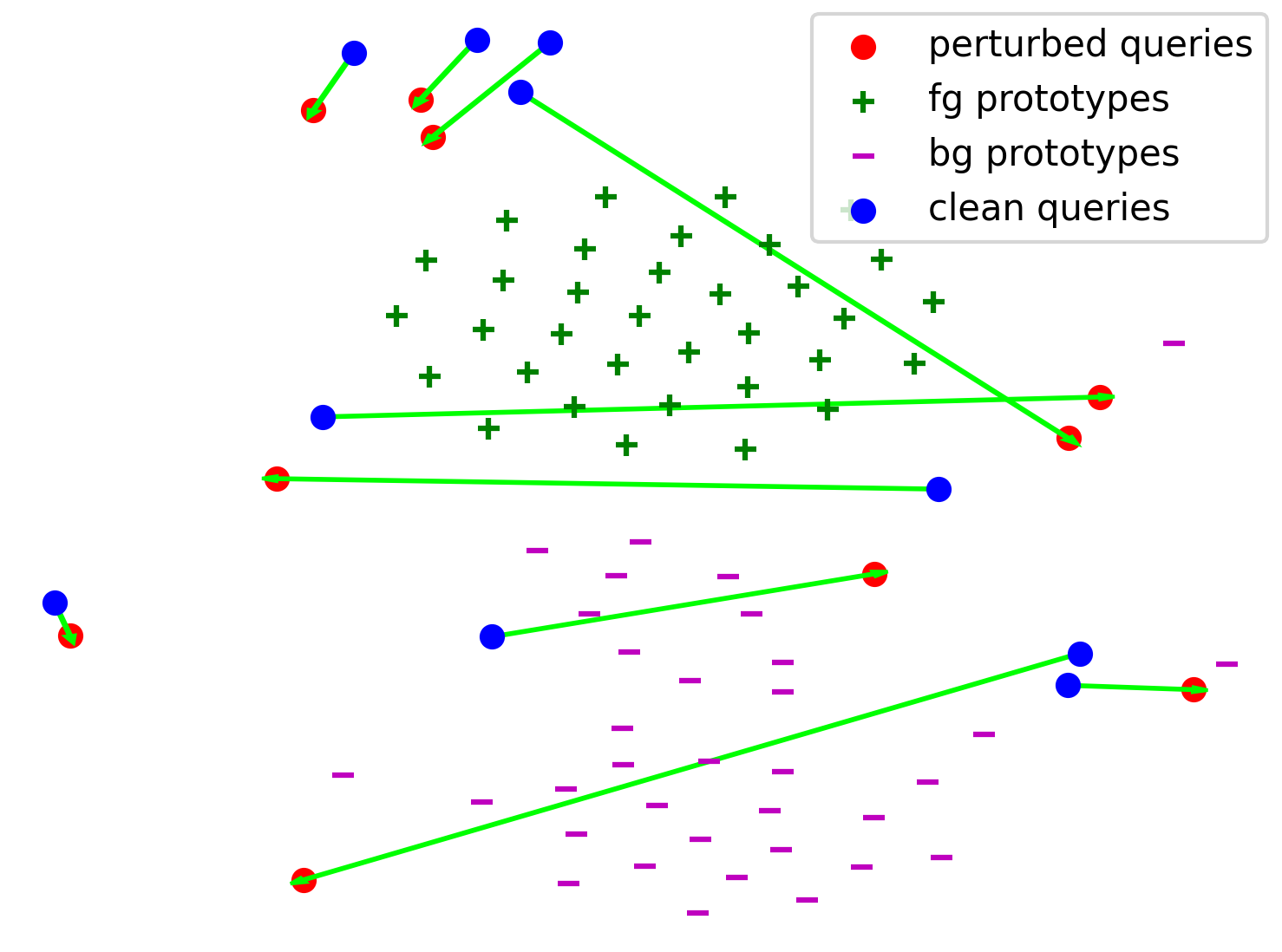}}
    \fcolorbox{black}{white}{\includegraphics[width=0.45\linewidth]{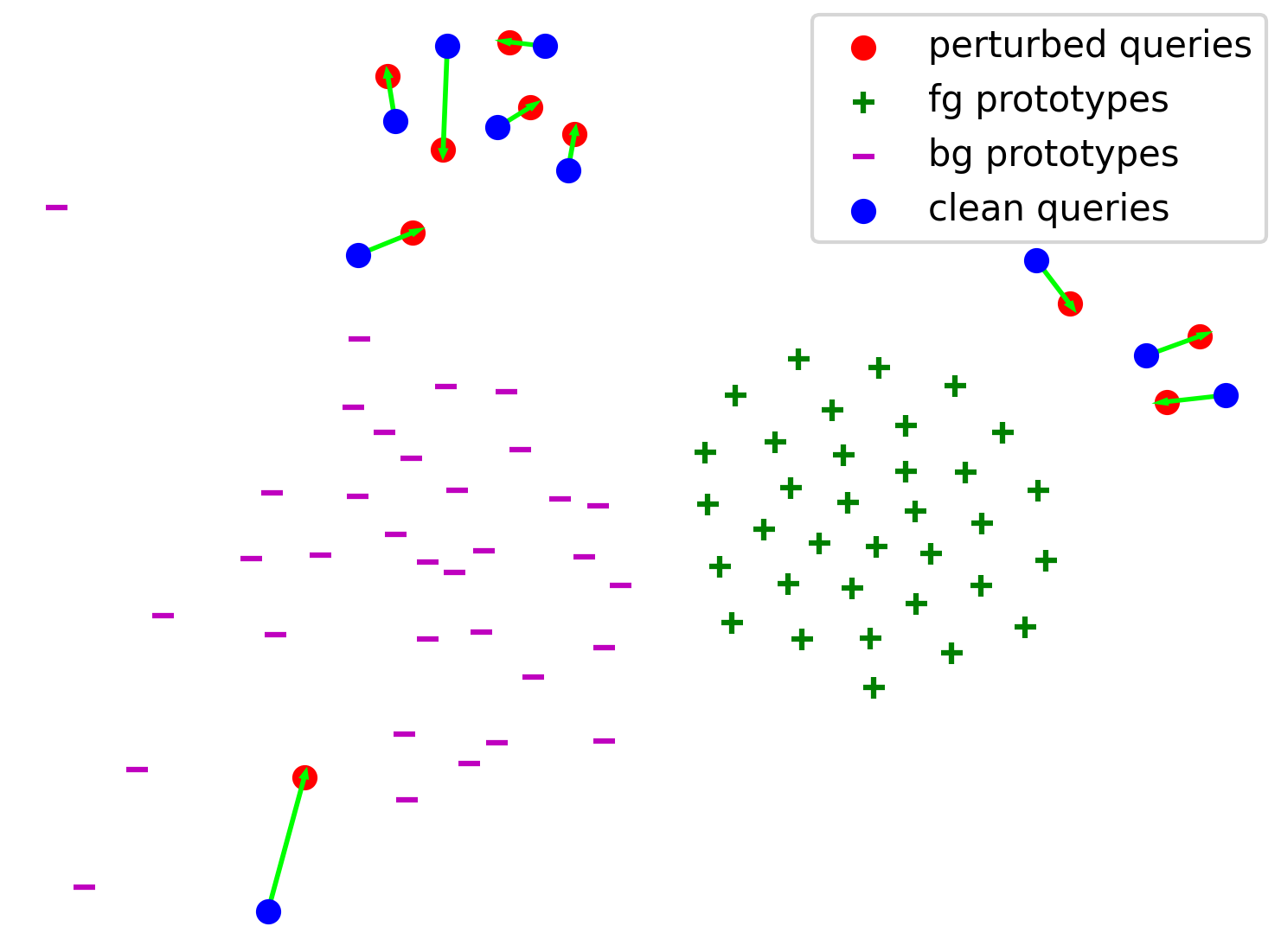}}
    \caption{t-SNE plots for query attack on AT-PANet (left) and R-PNODE (right). The clean query samples are connected to their corresponding perturbed samples using green arrows. The query pixels are sampled randomly from the set of available pixels across all query images.  \textcolor{black}{The prototypes are extracted across different episodes from the BCV dataset in the 1-way 1-shot in-domain Liver setting.}}
    \label{fig:tsne_q}
\end{figure}
\begin{figure}
    \centering
    \includegraphics[width=0.99\linewidth]{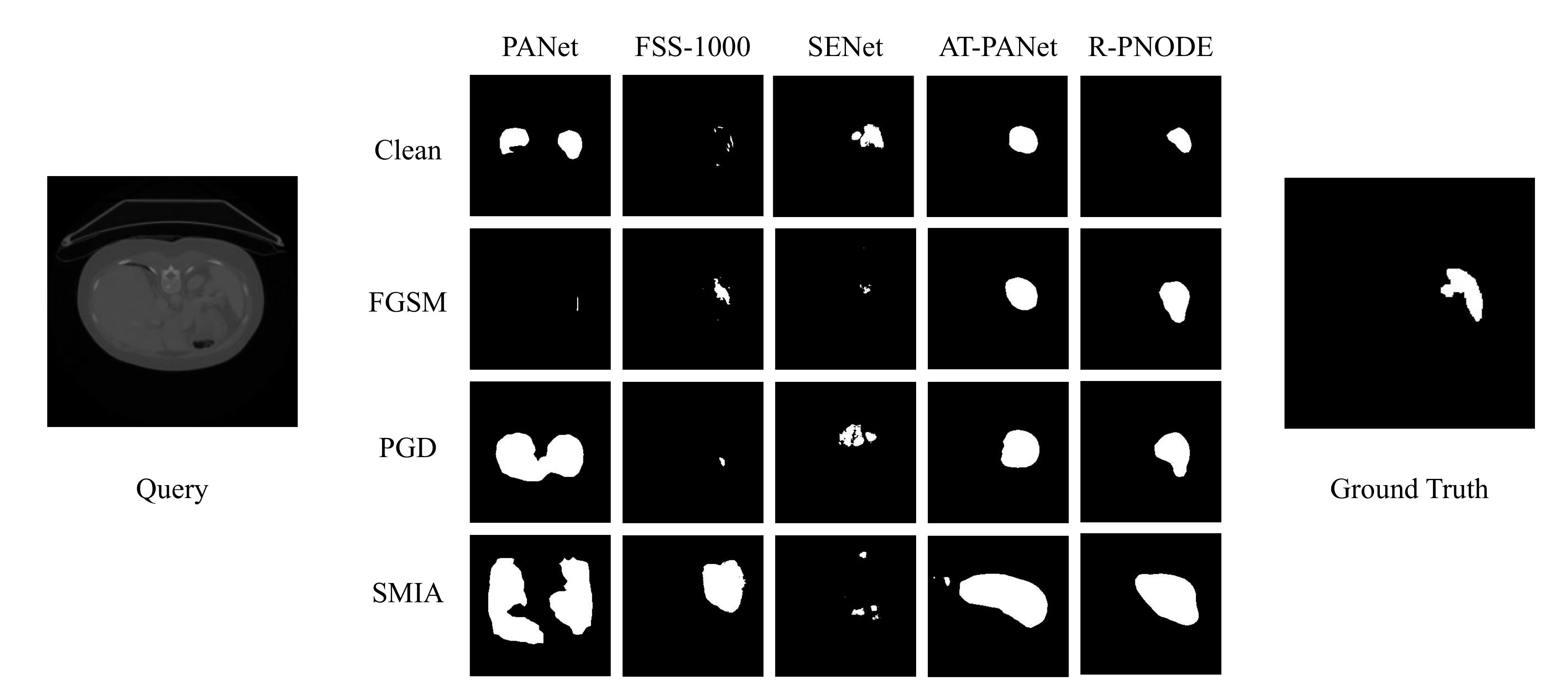}
    \caption{Predicted masks by different models for different attacks (attack hyperparameters are the same as those used in the tables). On the left is the query sample being tested and on the right is its corresponding ground-truth label for BCV $\rightarrow$ BCV in-domain setting for the Spleen organ.}
    % , BCV $\rightarrow$ CT-ORG (middle) and BCV $\rightarrow$ Decathlon (bottom)
    \label{fig:grid}
\end{figure}

 
% \begin{figure}[!t]
%     \centering
%     \fboxsep=0.5mm
%     \fboxrule=1pt  
%     \fcolorbox{black}{white}{\includegraphics[width=0.46\linewidth]{tsne_atpanet_support.png}}
%     \fcolorbox{black}{white}{\includegraphics[width=0.46\linewidth]{tsne_rpnode_support.png}}
%     \caption{t-SNE plots for support attack on AT-PANet (left) and R-PNODE (right). Clean support samples are connected to their corresponding perturbed ones using green arrows.}
%     \vspace{-2mm}
%     \label{fig:tsne_s}
% \end{figure}
We visualise the features ($Z_Q(T)$ and $Z_S(T)$) learnt by R-PNODE and the baseline AT-PANet using t-SNE plots in Fig.~\ref{fig:tsne_q} for query attacks. Here, the d-dimensional support prototypes are directly visualised, but for the query, the features of each pixel of an image are shown separately. It can be seen that while both models have query pixels that are significantly perturbed, the number of such perturbations is much smaller in the case of R-PNODE. Also, here most query pixels stayed closer to the same prototype cluster after perturbation than they were before. \textit{Consequently, the perturbation would not cause their misclassification.} 
%Similar observations can be made from Fig~\ref{fig:tsne_s} on the support attacks. Of the many prototypes, the perturbed prototypes of only a few are shown for better clarity.
% A clear difference between the magnitude of perturbation can be seen in the plots for AT-PANet and R-PNODE. An interesting observation is that while the perturbed prototypes are in the same relative locations in the case of R-PNODE, AT-PANET has the perturbed prototypes in a different order. The clean prototypes are such that the foreground class is above the background in the figure, but perturbed foreground class is actually below the perturbed background class. This would likely lead to mis-classification of query pixels and consequently poor performance for AT-PANet.
% \begin{figure*}
%     \centering
%     \includegraphics[width=\linewidth]{supplementary/perturbation.pdf}
%     \caption{Visualisation of the perturbed query images along with the predicted masks for the different attacks for Liver organ.}
%     \label{fig:viz}
% \end{figure*}
We visualise the predictions by each of these models for the different attacks in Fig.~\ref{fig:grid}. It can be seen that the predictions by R-PNODE are closest to the ground-truth labels. For the clean  samples, SENet, AT-PANet and R-PNODE are visually very similar, while FSS-1000 performs quite poorly and PANet predicts an extra object.  For FGSM,  AT-PANet performs much better than PANet, most likely because of encountering similar data during training. For PGD, only AT-PANet and R-PNODE are able to predict a mask that even resembles the ground truth. While FSS-1000 predicts poorly for clean, FGSM and PGD, it surprisingly performs well for SMIA along with R-PNODE, which also reflects in the dice scores in Table~\ref{tab:bcv_1shot_spleen}.
\section{Ablation Studies}
\textcolor{black}{We perform further experiments to understand the relative contributions of the various components in R-PNODE. We use the Spleen BCV in-domain setting to perform these experiments and report the results in Table~\ref{tab:ablations}. If the cluster and consistency losses are removed, the performance with different attacks drops by roughly \textcolor{black}{5-10\%}. 
%#COMMENTED
% This drop in performance, however is still not very high, when compared to the other baseline methods in Table~\ref{tab:bcv_1shot_spleen}. Thus, the model without the losses is itself adversarially robust as well. 
R-PNODE is inherently robust due to the intrinsic robustness provided by Neural-ODEs. The two losses then further improve robustness by regularizing it, leading to the improvements reflected here. Adding each loss gives some improvement in particular attacks and both losses together give the best overall performance. \textcolor{black}{They give an improvement of 6 absolute points on clean samples (from 58 to 64) on a scale of 0 to 100. Similarly, we get an improvement of 5 absolute points for FGSM (32 to 37), 4 absolute points for PGD (24 to 28) and 1 absolute point for SMIA (19 to 20). We also added these losses to PANet\cite{wang2019panet} and observed 1.6 and 7 absolute points improvement for Clean and PGD test queries, respectively over the baseline with the cluster loss. With consistency loss, we got 3 and 8 absolute points improvement for FGSM and PGD test queries, respectively over the baseline.
We perform an experiment where we removed the Neural-ODE block, gaussian perturbations and losses from R-PNODE that resulted in an FSS model with only a vanilla CNN block which we refer to as Vanilla CNN method in Table~\ref{tab:ablations}. As can be seen, there is a significant drop from R-PNODE’s original performance and thus, adding the Neural-ODE and losses helped improve the method’s performance.
}}
\begin{table}[!ht]
    \centering
    \setlength{\tabcolsep}{4pt}
    \renewcommand{\arraystretch}{1.2}
    \caption{Ablation studies of different components of R-PNODE for BCV $\rightarrow$ BCV in-domain for Spleen organ.}
    \scalebox{0.8}{
    \begin{tabular}{lcccc}
    \toprule
        Method & Clean & FGSM & PGD & SMIA \\
    \midrule
        % PNODE & &&& \\
        \textcolor{black}{Vanilla CNN} & 0.33 & 0.17 & 0.13 & 0.00\\
         \textcolor{black}{\textbf{R-PNODE}} & \textbf{0.64} & \textbf{0.37} & 0.28 & \textbf{0.20}\\
        without $\mathcal{L}_{CON}$ and  $\mathcal{L}_{CL}$ \textcolor{black}{(R-PNODE - $\mathcal{L}_{CON}$ - $\mathcal{L}_{CL}$)} & 0.58 & 0.32 & 0.24 & 0.19\\
        without $\mathcal{L}_{CL}$ & 0.58 & 0.30 & 0.26 & 0.19\\
        without $\mathcal{L}_{CON}$ & 0.59 & 0.34 & 0.25 & 0.19\\
        additive gaussian noise & 0.56 & 0.28 & 0.25 & 0.17 \\
        multiplicative gaussian noise $\sigma$ = 0.05 & 0.61 & 0.31 &  0.21 & 0.17\\
        multiplicative gaussian noise $\sigma$ = 0.3 & 0.60 &  0.30 & 0.27 & 0.15 \\
        multiplicative gaussian noise $\sigma$ = 0.5 & 0.59 &  0.29 & \textbf{0.30} & 0.14 \\
    \bottomrule
    \end{tabular}}
    \label{tab:ablations}
\end{table}

\textcolor{black}{While generating the gaussian samples, if additive gaussian noise is used instead of multiplicative, the performance drops severely. The subtle difference between additive and multiplicative noise lies in the fact that in the additive case, images of different scales are perturbed by the same absolute magnitude, while in the  multiplicative case, the perturbation has a relative magnitude. This may lead to lower performance since an image with a relatively smaller pixel intensity is perturbed by the same amount as another with a larger pixel intensity. Noise that may be ideal for the latter may be too much for the former. On the other hand, using a larger $\sigma$ ($\sigma \geq 0.3$) for multiplicative noise also leads to a slight decrease in performance. Thus, it may be inferred that an ideal choice of the type and strength of noise is important for the model's performance. While the overall theoretical utility may not change much, the subtle differences in its nature can become important.}
% \begin{figure}[!h]
%     \centering
%     \includegraphics[width=\linewidth]{eps.pdf}
%     \caption{Trend of performance of models on different intensities of FGSM query attack for BCV$\rightarrow$BCV in-domain Liver.}
%     \label{fig:plots}
% \end{figure}
% We also explore how the performance of different models changes with different attack intensities. We experiment on the BCV in-domain Liver setting with FGSM attacks for these experiments, and visualise the results in Fig~\ref{fig:plots}. As would be expected, the performance tends to drop with increasing attack intensity for all the models. An interesting observation is that the slope is somewhat higher for R-PNODE than it is for AT-PANet, even though they have very close performance for higher intensities. Thus, AT-PANet somewhat sacrifices the clean performance to get higher perturbed performance, while R-PNODE has a more natural trend  of greater performance boost with lesser attack intensity. 
\section{Conclusion}
% Few-Shot segmentation is a useful method to  segment novel regions in query images with the help of few support examples. 
\textcolor{black}{The existing FSS methods may misclassify query image pixels, even if the query image features are slightly perturbed. There is a lack of an explicit mechanism in these methods to constrain support and query image features of the same class to lie close in the representation space. Further,
defence against adversarial attacks on FSS models is of utmost importance as these models are data-scarce. 
% With their applications in the medical domain, it is critical to design methods that are robust against attacks of varying intensity and design. 
Although adversarial training may alleviate the risk associated with these attacks, the training procedure's computational overhead and poor generalizability render it less favourable, and sometimes impractical, as a robust defence strategy. 
We overcome limitations in existing FSS methods by employing Neural-ODEs to propose regularised and adversarially robust prototypical FSS method R-PNODE that stabilizes the model against various support and query perturbations. While the Neural-ODE block provides intrinsic robustness to the model, regularisation using cost-effective Gaussian samples further improves this robustness. With extensive experimentation, 
% on three openly available medical segmentation datasets, 
R-PNODE is shown to have better generalization abilities and adversarial robustness. To the best of our knowledge, we are the first to study the effects of different adversarial attacks on FSS models and provide a robust defence strategy.}
% that we hope will help the medical community.}
\bibliographystyle{IEEEtran}
\textcolor{black}{
\bibliography{bibtex}
}
\end{document}

%% file: clean.tex
\begin{table*}[!ht]
    \centering
    \setlength{\tabcolsep}{10pt}
    \renewcommand{\arraystretch}{1.2}
    \caption{Few-Shot organ segmentation results on 1-shot setting for BCV in-domain and cross-domain (BCV $\rightarrow$ CT-ORG and BCV $\rightarrow$ Decathlon) settings for Liver and Spleen organs (novel classes). The dice scores are rounded off to two decimals.}

    \scalebox{0.82}{
    \begin{tabular}{cc|cc|c|cc}
    \toprule
        \multirow{2}{*}{Method} & \multirow{2}{*}{Venue}& \multicolumn{2}{c|}{BCV $\rightarrow$ BCV} & \multicolumn{1}{c|}{BCV $\rightarrow$ CT-ORG} & \multicolumn{2}{c}{BCV $\rightarrow$ Decathlon}\\ 
    % \midrule
         &  & Liver & Spleen & Liver & Liver & Spleen \\
    \midrule
        PANet\cite{wang2019panet} & CVPR'19 & 0.61 ± 0.01 & 0.38 ± 0.03 & 0.52 ± 0.01 & 0.53 ± 0.01 & 0.43 ± 0.04\\
        FSS-1000\cite{li2020fss} &  CVPR'20 & 0.37 ± 0.04 & 0.41 ± 0.02 & 0.29 ± 0.04 & 0.37 ± 0.06  & 0.39 ± 0.01\\
        SENet\cite{roy2020squeeze} & MedIA'20 & 0.61 ± 0.01 & 0.57 ± 0.01 & 0.47 ± 0.01 & 0.50 ± 0.01 & 0.53 ± 0.01\\
        BiGRU\cite{kim2021bidirectional} & AAAI'21 & 0.51 ± 0.01 & 0.56 ± 0.01 & 0.43 ± 0.05& 0.46 ± 0.05& 0.46 ± 0.05\\
        SONet\cite{huang2021adversarial} & MSML'22 & 0.56 ± 0.00& 0.32 ± 0.00 & 0.28 ± 0.00 & 0.51 ± 0.00& 0.31 ± 0.00\\
        AT-PANet & - & 0.65 ± 0.01 & 0.46 ± 0.01  & 0.56 ± 0.01 & 0.57 ± 0.01 & 0.45 ± 0.01\\
        \textbf{R-PNODE} & - & \textbf{0.79 ± 0.01} & \textbf{0.64 ± 0.02} & \textbf{0.67 ± 0.01} & \textbf{0.79 ± 0.01} & \textbf{0.57 ± 0.01}\\
    \bottomrule
    
    \end{tabular}}

    \label{tab:clean}
\end{table*}

\begin{figure*}
    \centering
    \includegraphics[width=0.329\linewidth]{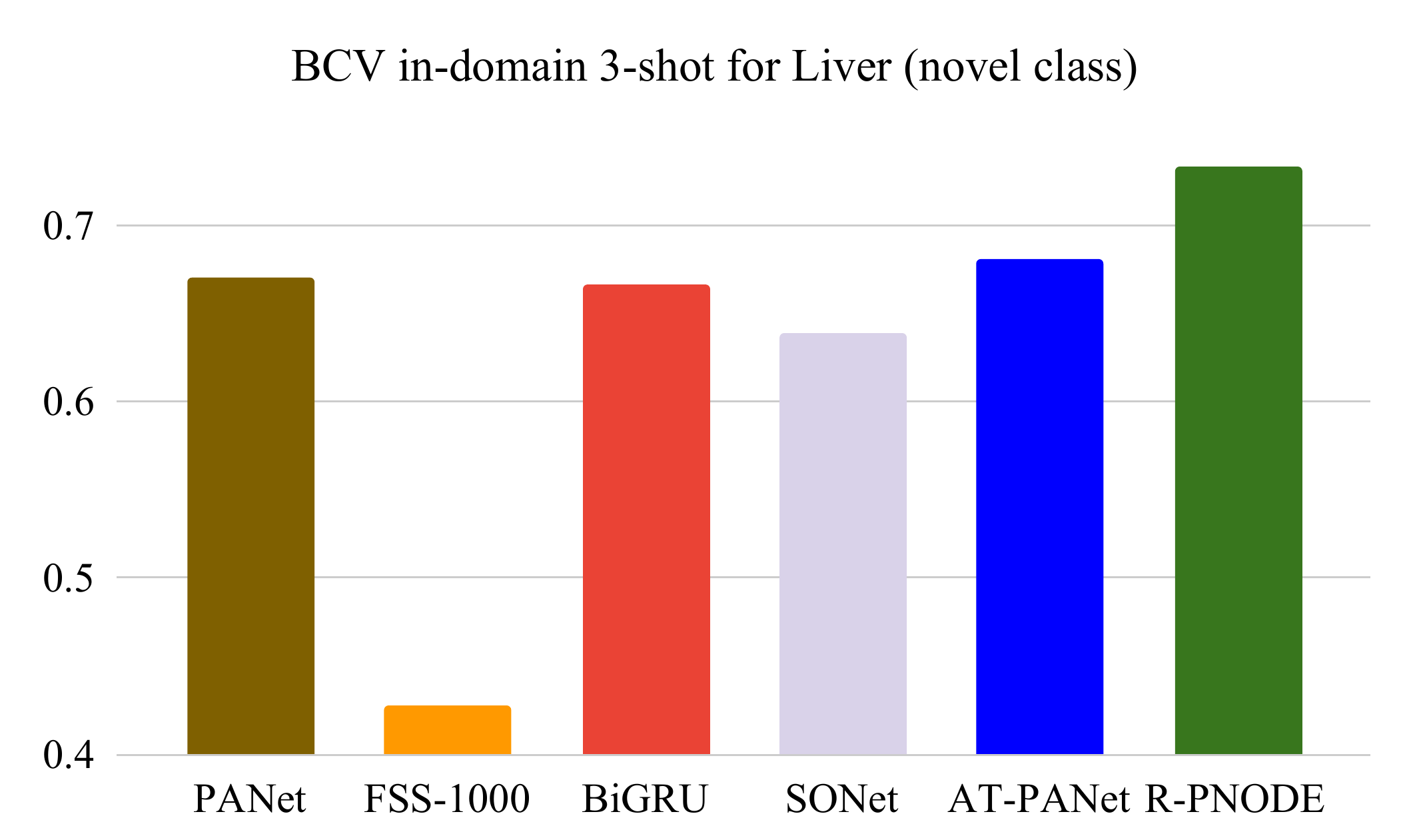}
    \includegraphics[width=0.329\linewidth]{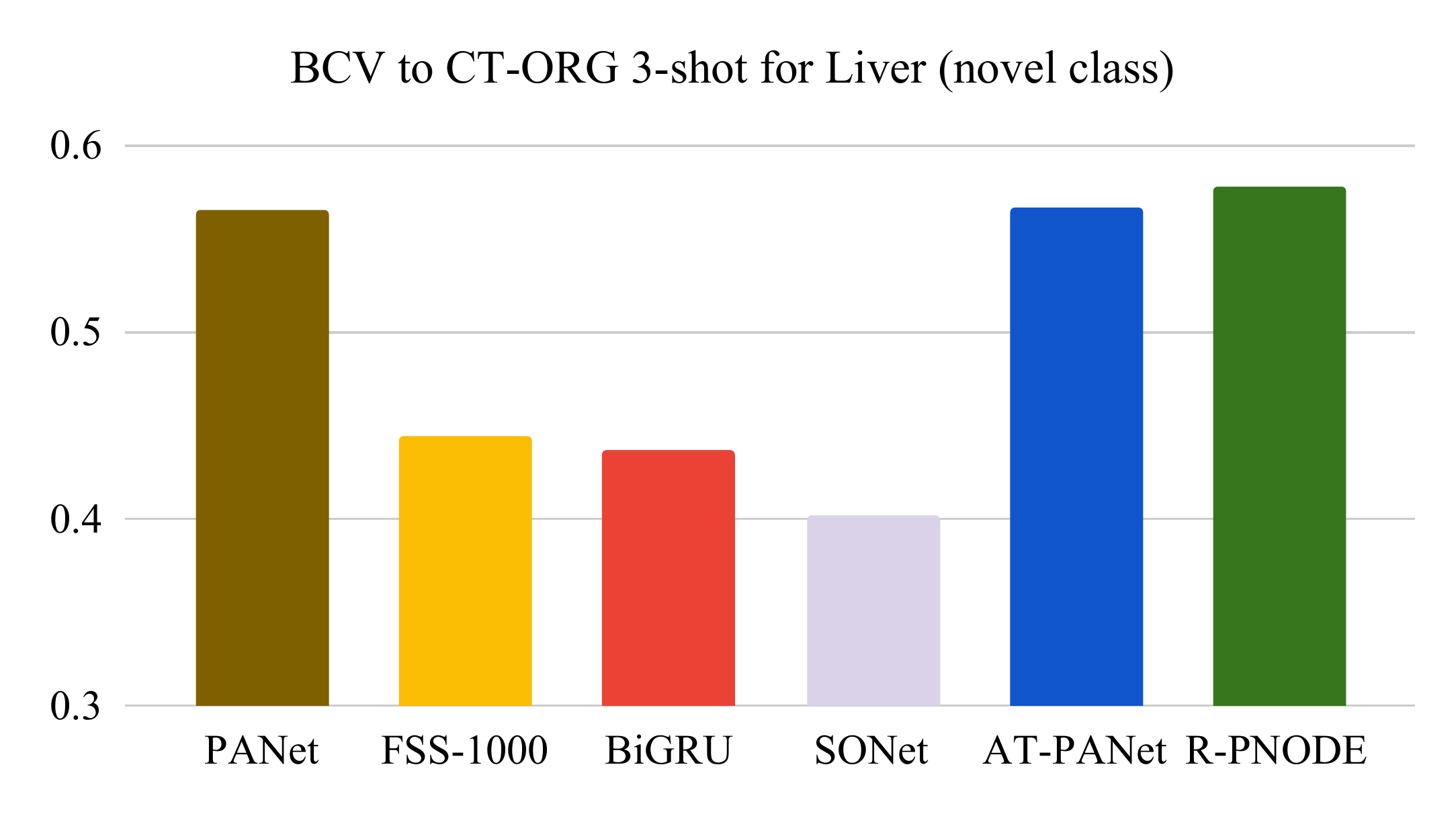}
    \includegraphics[width=0.329\linewidth]{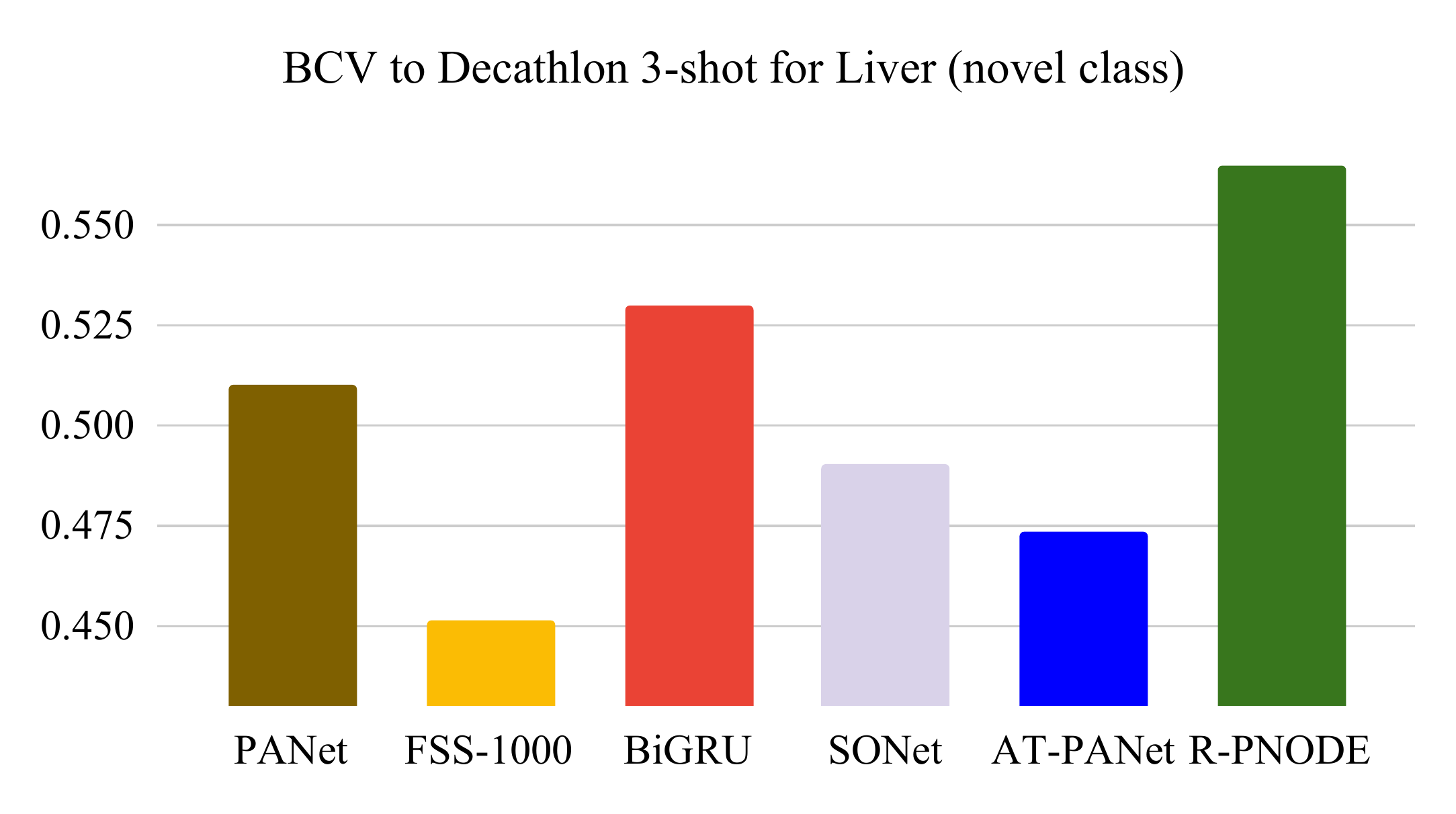}
    \caption{3-shot FSS results for BCV $\rightarrow$ BCV (in-domain), BCV $\rightarrow$ CT-ORG and BCV $\rightarrow$ Decathlon for Liver organ (novel class).}
    \label{fig:3shot_clean}
\end{figure*}